\documentclass[11pt, a4paper, logo]{lumia}

\usepackage[sort&compress]{natbib}
\usepackage{latexsym}
\usepackage{multirow}
\usepackage{array}
\usepackage{xspace}
\usepackage{makecell}
\usepackage{placeins}
\usepackage{adjustbox}
\usepackage{float}
\usepackage{wrapfig}

\definecolor{abstractbg}{HTML}{F2F5F8}
\AtBeginEnvironment{quote}{\raggedright}

\newcommand{\ii}{\ensuremath{I}\xspace}
\newcommand{\qthree}{Qwen3-8B\xspace}
\newcommand{\qthirty}{Qwen3-32B\xspace}
\newcommand{\llama}{Llama-3.1-8B\xspace}
\newcommand{\qtwo}{Qwen2.5-7B\xspace}
\newcommand{\haiku}{Claude Haiku 4.5\xspace}
\newcommand{\sonnet}{Claude Sonnet 5\xspace}

\hypersetup{pdftitle={Judged Useless, Queried Anyway: Tool-Using Agents Rarely Turn Their Own Evidence Judgments into Stopping Decisions}, pdfauthor={Chubin Zhang, Zhenglin Wan, Xingrui Yu, Jingxuan Wu, Yaxin Zhou, Ivor Tsang, Bo An}}
\setheadertitle{Judged Useless, Queried Anyway}
\title{Judged Useless, Queried Anyway: Tool-Using Agents Rarely Turn Their Own Evidence Judgments into Stopping Decisions}
\correspondingemail{\emailicon~\href{mailto:yu_xingrui@a-star.edu.sg}{yu\_xingrui@a-star.edu.sg} \quad $^\ddagger$ Corresponding author.}
\githublink{https://github.com/bennidict23/judged-useless-queried-anyway}
\makeatletter
\renewcommand{\lumia@githublink}{{\linkfont\faGithub\ Code: \href{\the\githublink}{github.com/bennidict23/judged-useless-queried-anyway}\\}}
\makeatother
\author{%
Chubin Zhang$^1$, Zhenglin Wan$^2$, Xingrui Yu$^{3,4\ddagger}$, Jingxuan Wu$^5$, Yaxin Zhou$^6$, Ivor Tsang$^{1,3,4}$, Bo An$^1$\\
$^1$ Nanyang Technological University, Singapore\\
$^2$ National University of Singapore, Singapore\\
$^3$ CFAR, Agency for Science, Technology and Research, Singapore\\
$^4$ IHPC, Agency for Science, Technology and Research, Singapore\\
$^5$ Department of Statistics and Operations Research, UNC-Chapel Hill, United States\\
$^6$ Carnegie Mellon University, United States
}

\begin{document}
\begin{abstract}
An agent whose tool keeps returning nothing useful should stop relying on it. In a retrieval environment with controlled source failures, we separate how agents judge results from what they do. We compare stopping at the same step after longer and shorter runs of results the agent judged useless; this contrast is zero for clock- or deadline-driven stopping. Where we record their judgments, the seven agents we test call a failing source's results useless 97--100\% of the time, yet most of them rarely stop on that judgment. Prompt cues change \emph{when} they stop but not \emph{what} they stop on. Permission to answer from memory and a reasoning mode can bring early stops regardless of evidence, a stated budget moves the 7--8B models' stops to the deadline, and a stopping rule or call cost in the prompt is followed at most partly. Stopping follows the evidence only when the harness enforces an integration step that makes the agent answer after five consecutive results it judged useless. This step raises failing-source success for every model, keeps the stopping point fixed when the budget doubles, and needs no extra judgment call when the agent states its judgments. A pre-registered replication on 300 fresh questions confirms the dissociation and the rule's effect.
\end{abstract}

\maketitle

\section{Introduction}

\begin{figure*}[t]
\centering
\begin{minipage}[t]{0.49\textwidth}\vspace{0pt}
\includegraphics[width=\linewidth]{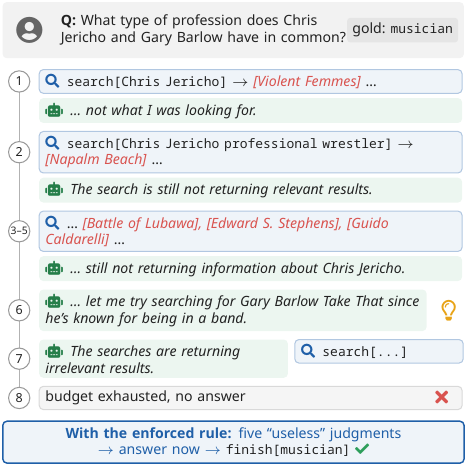}
\end{minipage}\hfill
\begin{minipage}[t]{0.45\textwidth}\vspace{0pt}
\centering
\includegraphics[width=\linewidth]{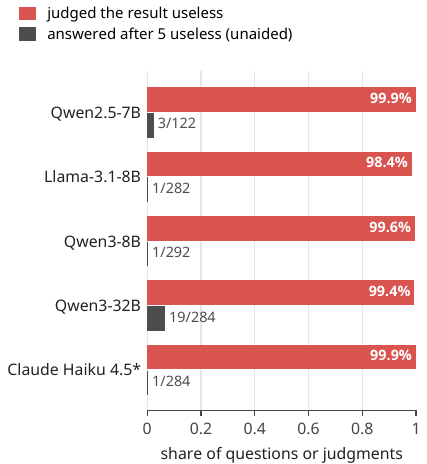}
\end{minipage}
\caption{\textbf{Left:} a real held-out trajectory of \haiku with a persistently failing source (selected by a fixed rule, Appendix~\ref{app:fig1}). It calls every result irrelevant and recalls at step~6 that Barlow is in a band, yet searches until its budget runs out. With the enforced rule, it answers correctly. \textbf{Right:} share of side-channel judgments that call a failing source's results useless (red), and share of questions with five such judgments in a row on which the unaided agent answers (gray). Judgments are replayed on the unaided agent's own trajectories, except for \haiku\ (*), whose judgments come from its enforced-rule run on the same questions.}
\label{fig:teaser}
\end{figure*}

Tools often fail silently. When a search index goes stale or a retriever drifts off topic, the agent still receives text, but the text no longer helps. Figure~\ref{fig:teaser} shows what a capable agent does then. In the example, \haiku tries to answer a two-hop question with a search tool that has stopped working. It calls every result irrelevant and recalls at step~6 that Barlow is in a band, yet it keeps searching until its budget runs out. Most agents we test behave this way. After their own judgments have called five results in a row useless, five of the seven answer on at most 7\% of questions. Yet the answer is often within reach. When forced to answer at that point, \haiku is right on 114 of the 284 questions where this happens, whereas on its own it answers only one of them.

Leaving a source is a sequential decision. One useless result is weak evidence that the source has failed, because a reformulated query may succeed and the source may recover. An agent that leaves at the first useless result forgoes the evidence of every source that would have recovered; one that never leaves spends its whole budget on every source that would not. A good policy must therefore \emph{integrate} its judgments: the longer the current run of results it has judged useless, the more likely it should be to stop. It should track the run rather than the total, because the run restarts when the source recovers.

Prior work documents that search-augmented models over-search \citep{xie2026oversearchingsearchaugmentedlargelanguage} and that deep-research agents keep browsing results that have turned out irrelevant, which \citet{zhang2026inertiaobjectivityimprovingdeep} attribute to distorted judgment. Two questions remain open. Does the failure lie in judging the evidence, in believing that continuing pays, or in turning either into an action? And can it be fixed simply by telling the agent more, for instance that it may answer without evidence or how many steps remain?

To answer them, we build a retrieval environment in which we control when the source fails and whether it recovers (\S\ref{sec:setting}), and we measure the agent's judgments, beliefs and actions separately. Success alone cannot reveal what stopping responds to. An agent that stops at a fixed step, one that stops at the deadline and one that stops after enough useless evidence can score alike and differ only when conditions change. The difference still matters, because an agent that waits for the deadline spends every call up to it and moves its stopping point whenever the budget changes. We therefore start from what integration predicts. If an agent integrates its own judgments, then \emph{at the same step} it should stop more often after an unbroken run of results it judged useless than after a history containing a result it judged useful (Figure~\ref{fig:dschema}). We test this prediction with the time-matched contrast $\Delta$, which is positive for an agent that integrates, zero for one that stops by the clock or the deadline, and negative for one whose stops follow useful evidence instead. Our main findings are:
\begin{itemize}[leftmargin=1.2em,itemsep=1pt,topsep=2pt]
\item \textbf{Agents judge correctly but do not act on their judgments} (\S\ref{sec:layers}). Every model whose judgments we recorded calls a failing source's results useless 97--100\% of the time, and the judgments the open models state in their own reasoning largely agree. Yet their stopping ignores these judgments. Every unaided open model has a negative $\Delta$, and the 7--8B models rarely answer even where their own beliefs favor answering.
\item \textbf{Telling agents more changes when they stop but not what they stop on} (\S\ref{sec:drivers}). Permission to answer from memory and a reasoning mode (\S\ref{sec:scale}) can bring early stops regardless of the evidence. A stated budget moves the 7--8B models' stops to the deadline, which shifts whenever the budget does. A stopping rule or call cost written into the prompt is followed at most partly, and showing agents their running count of useless judgments does not make them stop on it.
\item \textbf{An enforced integration step makes stopping follow the evidence} (\S\ref{sec:integration}). We let the harness apply the rule, so that only the \texttt{finish} action remains after five consecutive results the agent judged useless. Success on a failing source then rises for every model, and the stopping point stays put when the budget doubles. For agents that state their judgments in their reasoning, the harness can read the judgments there instead of asking for them.
\item \textbf{The pattern replicates and is not specific to 7--8B models} (\S\ref{sec:fever}, \S\ref{sec:scale}). It holds with $\Delta$ as the pre-registered primary endpoint on 300 fresh questions and transfers to fact verification. A larger open model, a reasoning mode and an RL-trained search agent still largely fail to stop on the evidence, and only \sonnet partly does so without being prompted.
\end{itemize}

\section{Setting}\label{sec:setting}

To study this decision, we need to know at every step whether the latest result helped and whether the source will recover. Real tools reveal neither, so we inject failures into question answering over small knowledge bases in which both are known.

\paragraph{Tasks and tools.} Questions come from the HotpotQA distractor development set \citep{yang2018hotpotqa}. Each question has its own knowledge base of the ten paragraphs HotpotQA provides. Two of them hold the supporting facts, so we know whether each result is useful. The agent follows the ReAct format \citep{yao2023react} with \texttt{search} (title match, then word overlap), \texttt{lookup} (next sentence with a keyword) and \texttt{finish}. A budget of eight actions is enough to answer a two-hop question after several failed searches. An answer is correct at token F1 $\ge 0.6$. The base prompt states neither the budget nor that an unanswered question counts as wrong. Because this alone could make continued search reasonable, \S\ref{sec:drivers} tests prompts that state both. For transfer, we use 292 balanced FEVER claims \citep{thorne2018fever}, where a random guess is right half the time. Each claim's knowledge base holds its evidence pages and eight from other claims.

\paragraph{Source failures.} A failed observation is replaced by a paragraph from another question's knowledge base. Such a paragraph is almost always useless, so we can score the agent's judgments. Section~\ref{sec:fever} tests harder, on-topic failures. Six regimes vary when failure happens, so that neither always leaving nor never leaving is right: \emph{clean} (no failure), \emph{persistent} (every observation fails), \emph{recover\_after\_$c$} (the first $c\in\{1,2,3\}$ fail, which rewards persistence) and \emph{late\_onset\_from\_3} (failure from the third observation on, which separates the length of a failure run from elapsed time). Success is averaged over the six with equal weight (\emph{mean6}).

\paragraph{Judgments, beliefs and actions.} At the decision point after the $t$-th result, we record three things. The first is the agent's \emph{judgment} of the latest result. Agents often state it in their next thought, though not always explicitly. To obtain a judgment at every step without changing the trajectory, we also ask a one-word question (USEFUL or USELESS) on a scratch copy of the conversation that the agent never sees again (the \emph{side channel}). When a condition acts on these judgments, they are recorded during the run; otherwise they are collected afterwards by replaying the recorded trajectories (\haiku's come from its enforced-rule run). The second is its \emph{beliefs}. On another scratch copy, the agent reports its probability of success if it continues and if it answers now. They let us test whether the agent searches on because it thinks continuing pays, and we compare them with the realized success of the recorded trajectory and of an answer-now branch replayed from the same history. The third is its \emph{action}.

\begin{wrapfigure}{r}{0.46\textwidth}
\centering
\includegraphics[width=\linewidth]{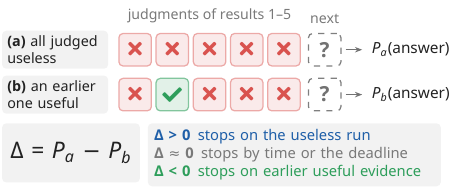}
\caption{What $\Delta$ compares at a fixed step. The enforced rule answers in (a) and continues in (b).}
\label{fig:dschema}
\vspace{-0.9\baselineskip}
\end{wrapfigure}
\paragraph{Measuring what stopping responds to.} If an agent integrates its judgments, its probability of answering should rise with the length of the current run of results it has judged useless. The obvious test is our pre-registered integration index $\ii = h(\{4,5\}) - h(\{1,2\})$, which compares the probability $h$ of answering after four or five consecutive failed results with that after one or two. But when the source fails from the start, the run grows with time. An agent that merely stops later also scores $\ii>0$, and a simulated clock-driven agent reaches $+0.5$. We therefore compare histories \emph{at the same step} (Figure~\ref{fig:dschema}). The time-matched contrast $\Delta$ is the probability of answering when every result so far was judged useless (a), minus that when the latest result was judged useless but an earlier one useful (b). We pool it over decisions after three to six results with the Mantel--Haenszel risk difference and exclude the final action, because it is the last chance to answer (Appendix~\ref{app:add12}).

\begin{wraptable}{r}{0.5\textwidth}
\centering\footnotesize\setlength{\tabcolsep}{3pt}\renewcommand{\arraystretch}{1.12}
\begin{tabular*}{\linewidth}{@{\extracolsep{\fill}}l l l@{}}
\toprule
Condition & What changes & Who stops \\
\midrule
none & base prompt & agent \\
permit & may answer from memory & agent \\
budget & permit + budget, counter & agent \\
stated rule & permit + rule in words & agent \\
call cost & budget + price per call & agent \\
decide & running count shown & agent \\
enforced rule & only \texttt{finish} after the run & harness \\
combo & budget + enforced rule & harness \\
\midrule
Measure & Based on & Role \\
\midrule
$\Delta$ & own judgments & primary \\
\makecell[l]{design-label\\contrast} & failure onset (design) & fallback \\
integration index $\ii$ & failure run (design) & confounded \\
hazard slope & own judgments & exploratory \\
\bottomrule
\end{tabular*}
\caption{Conditions (top) and stopping measures (bottom). Both the stated and the enforced rule prescribe answering after five consecutive results judged useless. $\Delta$ and $\ii$ are pre-registered; prompts in Appendix~\ref{app:prompts}.}
\label{tab:conditions}
\vspace{-1.8\baselineskip}
\end{wraptable}
\paragraph{Reading $\Delta$.} History (b) is the natural comparison because its latest result was also judged useless, so the two histories differ in how much useless evidence has accumulated. An agent that holds useful evidence may also reasonably answer, which pushes $\Delta$ down. Hence $\Delta$ is zero for stopping by time or the deadline, positive for stopping on the useless run, and negative when stopping follows earlier useful evidence (Figure~\ref{fig:dschema}). A hazard model that separates the useless run from earlier useful evidence confirms this reading (\S\ref{sec:layers}), and on simulated agents $\Delta$ separates time-, deadline- and evidence-driven policies as intended (Appendix~\ref{app:add12}). Where judgments are unavailable, the \emph{design-label contrast} makes the same comparison between a source failing from the start and one failing from the third result.

\paragraph{Conditions.} Each condition adds one element to a simpler one (Table~\ref{tab:conditions}). Four tell the agent more: permission, the budget (eight actions, a counter, unanswered counts as wrong), a price of 0.05 per call against 1 per correct answer, or the stopping rule in words (\emph{stated rule}). The \emph{decide} condition shows the agent its running count but leaves the choice to it. The \emph{enforced rule} instead lets the harness act after $k=5$ consecutive useless judgments ($k$ set on development data). Two controls replace these judgments with random or lexical signals.

\paragraph{Models and protocol.} We run \qtwo, \llama, \qthree and \qthirty \citep{qwen2025qwen25technicalreport,grattafiori2024llama3herdmodels,yang2025qwen3technicalreport} locally from their official weights with vLLM \citep{kwon2023efficientmemorymanagementlarge}. All four run without thinking, using the Qwen3 models' recommended sampling and greedy decoding for the others. Section~\ref{sec:scale} adds a reasoning mode and a search agent trained with reinforcement learning. To limit cost, the Claude models run through the official API in fewer conditions. Both run unaided and with the enforced rule (\sonnet on 100 questions), and \haiku also runs with the rule or the cost stated on 200 questions. Every threshold was set on 100 development questions; confirmatory results use a disjoint set of 300 (test300) and the FEVER claims. All hypotheses were registered before the corresponding held-out runs. The amendments that define $\Delta$ followed earlier results on the same questions, so we re-test their predictions on 300 questions never used before (fresh300). Appendix Table~\ref{tab:map} marks which claims rest on pre-registered tests and which on exploratory analyses. Results are raw unless marked repaired, and Appendix~\ref{app:prereg} lists the deviations. Differences are paired by question with 2,000 bootstrap resamples, and tests across models are Holm-corrected.

\paragraph{Measurement validity.} Stated judgments have to be read from free text, so two LLM annotators from different model families labeled them, with author adjudication (binary $\kappa$ 0.90; Appendix~\ref{app:annot}). All other results rest on design labels, side-channel judgments or deterministic action parsing, which had no errors in a 100-point audit. Some answers copied the template \texttt{finish[answer]}; we repaired them with one fixed procedure, and the main conclusions do not change (Appendix~\ref{app:repair}).

\section{Agents judge correctly but do not act on their judgments}\label{sec:layers}

Continued search would be reasonable if the agent misjudged the results or believed that continuing pays. We test both explanations and then ask whether actions follow the judgments.

\paragraph{Judgments are accurate.} Agents do recognize useless results. Through the side channel, every model whose judgments we recorded calls a failing source's results useless 97--100\% of the time on HotpotQA (Figure~\ref{fig:teaser}) and FEVER, and the judgments \qthree states in its own reasoning are about as accurate (Appendix~\ref{app:annot}). This is not a bias toward ``useless'', since models call a gold paragraph useful 77--98\% of the time when it is first retrieved, except \qtwo\ (57\%; Appendix~\ref{app:spec}).

\paragraph{Beliefs alone do not explain it.} A second explanation is that agents know the results are useless but believe that continuing still pays. On a failing source, \qthree does overestimate the value of continuing, but \qtwo overestimates that of answering now and \llama underestimates it (descriptive probe, Appendix~\ref{app:beliefs}). Beliefs thus point in different directions across models, and actions do not follow them. Where their own reports favor answering now, \llama answers at only 32 of 595 decision points and \qtwo at 267 of 2,443 (persistent and plausible regimes; \S\ref{sec:fever}).

\paragraph{Actions do not follow the judgments.} After judging five consecutive results useless, unaided \qthree answers on only 1 of 292 questions, and the other models in Figure~\ref{fig:teaser} answer on at most 7\%. Otherwise they search until the budget runs out. Stopping there would have paid, since answering at that point succeeds 15--17\% of the time and continuing at most 7\% (7--8B models, exploratory; Appendix~\ref{app:cap}). Stopping also does not become more likely as useless judgments accumulate, as the negative $\Delta$ of every open model shows (Figure~\ref{fig:delta}). A negative $\Delta$ could also mean answering after useful evidence, so we fit a discrete-time hazard model that separates the two (exploratory; Appendix~\ref{app:hazard}). In this model, each further useless judgment changes the unaided 7--8B agents' probability of answering by $-0.02$ to $-0.01$, against $+0.15$ to $+0.17$ under the enforced rule.

\paragraph{The same holds for judgments stated in reasoning.} The judgments the open models state in their reasoning agree with the side channel on 87--96\% of decision points. When $\Delta$ is computed from these stated judgments instead (Appendix~\ref{app:add16}), it stays below zero, and after five stated useless judgments in a row the agents answer on 0--0.4\% of questions (\qthirty: 6\%).

\section{Telling agents more changes when they stop but not what they stop on}\label{sec:drivers}

\begin{figure*}[t]
\centering
\includegraphics[width=\textwidth]{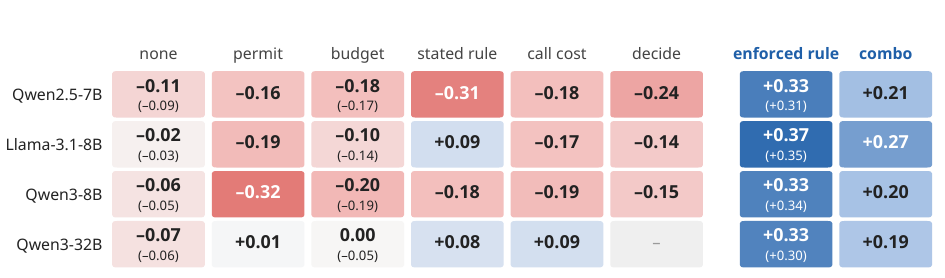}
\caption{Only when the harness stops (blue labels) does stopping follow the agent's judgments for every model. Cells show $\Delta$ from the agents' own judgments in test300 (blue $>0$, red $<0$), with the fresh300 replication in parentheses. Decide is exploratory; intervals are in Appendix~\ref{app:add12}.}
\label{fig:delta}
\end{figure*}

If agents keep searching because they lack some information, providing it should make them stop on the evidence. We test the candidates one at a time, mainly on the 7--8B models (Table~\ref{tab:main}, Figure~\ref{fig:delta}).

\begin{wraptable}{r}{0.46\textwidth}
\centering\footnotesize\setlength{\tabcolsep}{4pt}\renewcommand{\arraystretch}{1.12}
\begin{tabular*}{\linewidth}{@{\extracolsep{\fill}}l cccc@{}}
\toprule
Condition & \makecell{Qwen2.5\\7B} & \makecell{Llama-3.1\\8B} & \makecell{Qwen3\\8B} & \makecell{Qwen3\\32B} \\
\midrule
none & .192 & .263 & .448 & .509 \\
permit & .213 & .269 & .423 & .517 \\
budget & .328 & .312 & .464 & \textbf{.581} \\
stated rule & .206 & .267 & .439 & .526 \\
call cost & .326 & .317 & .458 & .575 \\
decide & .360 & .266 & .426 & -- \\
\midrule
enforced rule & .237 & .328 & .474 & .572 \\
combo & \textbf{.375} & \textbf{.337} & \textbf{.478} & .578 \\
\bottomrule
\end{tabular*}
\par\vspace{6pt}
\begin{tabular*}{\linewidth}{@{\extracolsep{\fill}}l cc@{}}
\toprule
Agent & none & enforced rule \\
\midrule
Claude Haiku 4.5 & .501 & \textbf{.598} \\
Claude Sonnet 5$^\dagger$ & .632 & \textbf{.653} \\
Search-R1$^\ddagger$ & .403 & \textbf{.427} \\
\bottomrule
\end{tabular*}
\caption{The budget condition matches or exceeds the enforced rule's success on several models, so success alone cannot show what stopping responds to. Mean6 success (test300, raw; bold: best per model; --: not run). Bottom: agents run only unaided and with the enforced rule ($^\dagger$first 100 questions; $^\ddagger$RL-trained, own prompt). Intervals and per-regime results: Appendix~\ref{app:tables}.}
\label{tab:main}
\vspace{-1.8\baselineskip}
\end{wraptable}
\paragraph{Permission yields indiscriminate stopping.} Perhaps agents believe they may not answer without evidence. Telling them they ``may also answer from [their] own knowledge'' leaves \llama and \qtwo almost unchanged. Permission does make \qthree stop earlier, but its stops ignore the evidence. The model answers before any real result on 55\% of questions whose source recovers after three failures, and its $\Delta$ drops to $-0.32$.

\begin{wrapfigure}{r}{0.46\textwidth}
\centering
\includegraphics[width=\linewidth]{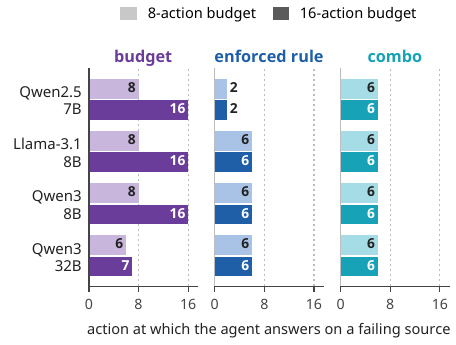}
\caption{Budget-cued 7--8B models answer at whatever the deadline is, whereas the enforced rule and the combination answer at the same action under both budgets. Bars: median action at which agents answer on a persistently failing source (among answered questions) with an 8-action (light) and a 16-action (dark) budget. \qtwo answers early under the enforced rule because even unaided it stops after one useless result.}
\label{fig:budget}
\vspace{-1.8\baselineskip}
\end{wrapfigure}

\paragraph{A stated budget yields stopping at the deadline.} Or agents may not know that steps are limited and that an unanswered question counts as wrong. Stating the budget with a counter does make the 7--8B models answer, but on the last action (51--79\% of answers on a failing source), and their $\Delta$ stays negative. This condition matches the enforced rule's success for \qthree and \llama and exceeds it for \qtwo, so success alone cannot tell the two apart. A change of deadline separates them. With the budget doubled, budget-cued agents move their answers from the eighth to the sixteenth action and roughly double their tool calls on a failing source without any gain in success (Figure~\ref{fig:budget}).

\paragraph{Stating the rule or the cost is not enough.} If agents simply did not know which stopping rule to use or what a call costs, writing these into the prompt should work. With the rule stated in words, only \llama and \qthirty follow it, and only partly ($\Delta=+0.09$ and $+0.08$, against $+0.37$ and $+0.33$ when enforced). \qtwo and \qthree answer early whatever they judged, and no model reaches the enforced rule's success. With a price of 0.05 per call added to the budget, the 7--8B models still answer at the deadline and do no better than with the budget alone; only \qthirty stops partly on its judgments. \haiku behaves similarly. With the rule stated, its stopping follows the evidence only partly. With the price stated, it answers on almost every question regardless of the evidence and even rivals the enforced rule's success (Appendix~\ref{app:add19}). Writing the rule or the price into the prompt thus supplies the integration step only partly.

\paragraph{A backup tool does not make leaving evidence-driven.} A different reason to stay is that leaving means giving up. We therefore added a second tool, \texttt{backup\_search}, that searches an intact copy of the knowledge base. The prompt describes it as slower and costlier and suggests it when \texttt{search} is not useful (Appendix~\ref{app:add13}). When the primary search fails throughout, \qthree and \qthirty reasonably switch at the first useless result. By the fourth action, however, \llama and \qtwo have switched on only 17\% and 49\% of questions, and \llama's success falls to 0.05 (0.50 with a working primary). Yet no unaided agent becomes more likely to leave as useless results accumulate ($\Delta\le+0.02$), whereas disabling the failing tool after five useless judgments makes leaving evidence-driven ($\Delta=+0.26$ and $+0.30$ for \llama and \qthree; Appendix~\ref{app:add13}).

\paragraph{Even the running count is not enough.} Agents might also judge each result correctly but lose count. The decide condition therefore writes the agent's judgment and its running count of useless judgments into the context at every step and asks it to weigh continuing against answering now. Agents still do not stop on the count, and their $\Delta$ is $-0.14$ to $-0.24$ for the three 7--8B models (exploratory; Figure~\ref{fig:delta}). Only \qtwo gains appreciably on mean6 and even beats the enforced rule, but it does so by stopping early indiscriminately.

\section{Enforced integration makes stopping follow the evidence}\label{sec:integration}

The agents judge correctly, and neither information nor the running count makes the 7--8B models stop on their judgments. This leaves the step that turns accumulated judgments into an action, which we test directly by letting the harness take it.

\paragraph{An enforced integration step anchors stopping to the evidence.} The enforced rule's positive $\Delta$ ($+0.33$ to $+0.40$) is expected by construction. The informative questions are whether this step alone raises success and whether it frees the stopping point from the deadline. Mean6 rises for the four open models and \haiku ($+0.026$ to $+0.097$, Holm-significant). Where the source recovers or never fails, the rule is non-inferior to the unaided agent in 18 of 20 tests (margin $-0.05$; \qthree fails narrowly twice). With the budget doubled, the enforced rule answers at the same action as before (Figure~\ref{fig:budget}).

\paragraph{The gain comes from acting on an accurate signal.} Against a random signal that says ``useless'' equally often but at random steps, the enforced rule wins by $0.011$ to $0.029$ (Holm-significant), so the timing of the signal matters. A cheap lexical detector that calls a result useless when it shares few content words with the question does about as well as the agent (exploratory).

\paragraph{Deadline and integration combine.} The budget prevents premature answers and the enforced rule supplies the evidence-driven stop, so together they should keep both benefits. The combination is indeed non-inferior to both components on every model and task (14 of 14 pre-registered tests), and it beats the enforced rule by $+0.138$ for \qtwo, whose main failure is answering too early.

\paragraph{How good is the stopping point?} Against the best fixed cap or run-length policy chosen in hindsight, unaided agents lose 6--13 points of success (exploratory; Appendix~\ref{app:cap}), and with a doubled budget the budget-cued 7--8B Qwen models make 61--71\% more calls for no gain. The enforced rule also makes three to five judgment calls per question, which would put it below the budget condition if they were charged like tool calls. These calls can be avoided by reading the judgment from the agent's own thought with a keyword reader, which keeps success within 0.01 of the enforced rule for \llama and the Qwen3 models. Combined with the budget statement, the reader beats the budget condition alone at 0.05 per call for \llama and \qthree (Appendix~\ref{app:add19}). The reader does not help \qtwo, which rarely states a judgment.

\section{The pattern replicates on fresh questions, a second task and harder failures}\label{sec:fever}

\paragraph{The predictions hold on fresh questions.} Because $\Delta$ was defined after earlier results were known, we re-tested its predictions on 300 fresh questions, with $\Delta$ registered as the primary endpoint. All nineteen tests pass (Figure~\ref{fig:delta}, in parentheses; Appendix~\ref{app:add16}). Again, $\Delta$ stays below zero for the unaided and budget-cued agents and reaches $+0.30$ to $+0.35$ under the enforced rule, which beats the unaided agent on mean6 ($+0.030$ to $+0.061$). After five useless judgments, the unaided 7--8B agents answer on at most 3\% of questions.

\paragraph{On fact verification the effects are larger.} On FEVER a random guess is right half the time, so an agent that stops and answers from memory gains more. With every parameter carried over unchanged, the pattern replicates. Unaided agents answer after five useless judgments on at most 3\% of claims, and the enforced rule raises mean6 by $+0.129$ to $+0.201$. The budget condition again answers mostly on the last action (69--96\%; F-H10 in Appendix~\ref{app:prereg}).

\paragraph{Harder failures weaken judgments but not the dissociation.} Unrelated paragraphs are easy to recognize, whereas real failures are often on topic. When the failed observations are the question's own distractor paragraphs (\emph{plausible}) or its own pages with every supporting fact removed (\emph{answerless}), judgments are less accurate (84--97\% and 76--97\% useless), but the dissociation remains. After five useless judgments, the unaided agents answer on 2--23\% and 5--21\% of questions. The enforced rule still raises success in both, although on the answerless source the budget condition does better (Appendices~\ref{app:add12}, \ref{app:add13}). Two further runs with new seeds and failure draws reproduce the main contrasts (Appendix~\ref{app:rep}).

\section{Larger, reasoning and search-trained agents rarely integrate either}\label{sec:scale}

The 7--8B models might lack the capacity to integrate, so we test a larger open model, a frontier model, a reasoning mode and a search-trained agent.

\paragraph{Left alone, larger models still rarely integrate.} \qthirty judges failing results as accurately as the smaller models, yet answers after five such judgments on only 19 of 284 questions. \sonnet is the only unaided model whose stopping partly follows the evidence (design-label contrast $+0.12$ [$+0.05, +0.18$], exploratory). Still, unaided \sonnet exhausted its budget on 45\% of the questions where its enforced rule fired. The enforced rule raises its success on a failing source from 0.40 to 0.51, although on 100 questions the gain over six regimes is not significant ($+0.022$, $p=0.051$).

\paragraph{A budget cue makes a larger model stop earlier, but mostly by time.} With its budget stated, \qthirty answers around the sixth action and matches the enforced rule's success. Its integration index rises to $+0.32$, so our pre-registered prediction that it would stay near zero failed. The time-matched contrast shows what drives this rise. At the same step, \qthirty answers as often after earlier evidence it judged useful as after a run it judged useless ($\Delta=0.00$ [$-0.06, +0.05$]), and with the budget doubled it answers a step later and makes 46\% more tool calls for no gain.

\paragraph{Training to search does not teach integration.} Training to search might teach an agent when to stop, but it does not for Search-R1 \citep{jin2025searchr1trainingllmsreason}, which was trained with PPO on HotpotQA and NQ and runs here with its own prompt. On a failing source it answers from memory after one or two searches on 27\% of questions but searches to the end of its budget on 45\%. At the same step, it answers no more often after a longer failure (design-label contrast $-0.03$ [$-0.07, +0.01$]). Because its judgments also call most real results useless (89\%), we do not interpret its $\Delta$; the enforced rule still raises its mean6 by $+0.024$ (Appendix~\ref{app:add16}).

\paragraph{A reasoning mode makes agents stop earlier regardless of the evidence.} Thinking might let an agent integrate within its reasoning. Instead, in thinking mode \qthirty answers from memory on 97\% of failing-source questions (11\% without thinking). Where the source recovers after two failures, it answers before any real result on 47\% of questions ($\Delta=+0.01$ [$-0.18, +0.17$]). \qthree in thinking mode likewise stops no more often after a longer failure. Some of \qthirty's stops may follow the evidence only when its budget is stated (design-label contrast $+0.32$, from few late decisions; Appendix~\ref{app:add16}).

\section{Related work}\label{sec:related}

\paragraph{Agents and unreliable tools.} Agents carry wrong tool output into their answers even when their reasoning notices the conflict \citep{yang2026agentstrusttoolsmuch}, mostly fail to diagnose recoverable tool faults \citep{tian2026functioncallingbenchmarkingtoolusing}, and can be hurt by misleading feedback \citep{ming2025helpfulagentmeetsdeceptive}. We ask instead whether they stop relying on a source that returns nothing useful.

\paragraph{When to stop searching.} Search-augmented models over-search \citep{xie2026oversearchingsearchaugmentedlargelanguage,liu2026diagnosingsearchbehaviorfailure}. Remedies train the search/answer boundary \citep{zhang2026searchsearchaligningdecision}, verify sufficiency \citep{roh2026haltverificationawarestoppingretrievalaugmented}, add a controller \citep{soudani2026deepresearchagentsstagnate} or guard against unsupported stops \citep{liu2026agentstopevidencecarryingtermination}. Adaptive retrieval decides when to retrieve \citep{asai2023selfraglearningretrievegenerate,jiang2023activeretrievalaugmentedgeneration} or corrects poor retrievals with a learned evaluator \citep{yan2024correctiveretrievalaugmentedgeneration}, and RL-trained search agents learn when to stop \citep{jin2025searchr1trainingllmsreason}. These methods build the decision into the system, whereas we measure whether agents act on their own judgment that a source keeps failing.

\paragraph{Abstention and recovery.} Benchmarks ask whether agents know when to abstain \citep{luo2026agenticabstentionagentsknow} or not to act \citep{liu2026agentabstainllmagentsknow}. \citet{luo2026agenticabstentionagentsknow} use a fixed search budget, which we find sets a deadline. \citet{chen2026retryswitchabstainlearning} learn fallback strategies with reinforcement learning, whereas our rule uses only the agent's own judgments.

\paragraph{Knowing and doing.} Models largely know what they know \citep{kadavath2022languagemodelsmostlyknow}. A gap between judging and doing has nonetheless been shown for the decision to call a tool at all \citep{li2026knowingactingbenchmarkingselfawareness}, and making the judgment explicit reduces it \citep{cheng2026modeladaptivetoolnecessityreveals}. The gap also appears in bandit tasks, where models state the right strategy but act greedily \citep{schmied2025llmsgreedyagentseffects}. In our setting, an explicit judgment does not lead to stopping even when its running count is shown.

\section{Discussion and conclusion}\label{sec:discussion}

Agents recognize useless results, but their beliefs, what they are told and the running count all fail to make the 7--8B models stop on them, apart from \llama's partial following of the stated rule. The missing piece is integration. In the spirit of sequential testing \citep{wald1945sequential}, the enforced rule adds a run-length test that maps the count to an action. The step matters even where success does not show its absence. With a known deadline and free calls, answering at the deadline does not reduce success, but it spends every call and shifts whenever the budget does. Agent designs should therefore make this step explicit rather than expect it from prompting. Outcome-reward reinforcement learning did not teach it to Search-R1 either (\S\ref{sec:scale}). The step is also cheap, because the harness can read the judgment in the agent's reasoning instead of asking for it. Evaluations should not infer evidence-driven stopping from success, which a deadline also yields. They should instead test whether stopping at a fixed step rises with the run of results the agent judged useless. An agent's stated judgment of its evidence does not guarantee that it acts on it, so the two must be measured separately.

\paragraph{Limitations.} We simulate failures by replacing observations in fixed knowledge bases. Real tools also fail by returning partial, stale or erroneous content, and whether the pattern holds under such natural failures is untested, although on-topic failures leave the dissociation intact (\S\ref{sec:fever}). Our sources recover after at most three failures, and the rule's threshold was chosen for that distribution. When recovery comes later, the rule can give up just before it (Appendix~\ref{app:add13}), so the threshold would need tuning for real sources. Most controls ran on open models, and the two Claude models received fewer conditions and questions. Judgments are elicited on a copy of the conversation and beliefs are self-reports, so we compare them only within a model. Stated judgments were labeled by LLM annotators with author adjudication rather than by independent human annotators. Answering from memory is not automatically safe, since parametric knowledge can be stale or wrong. Our results therefore argue for making the decision to abandon a source explicit and evaluating it, rather than for abandoning tools.

\bibliographystyle{plainnat}
\bibliography{refs}

\clearpage
\appendix
\renewcommand{\topfraction}{0.95}\renewcommand{\dbltopfraction}{0.95}\renewcommand{\bottomfraction}{0.6}
\renewcommand{\textfraction}{0.05}\renewcommand{\floatpagefraction}{0.8}\renewcommand{\dblfloatpagefraction}{0.8}
\setcounter{topnumber}{4}\setcounter{dbltopnumber}{4}\setcounter{bottomnumber}{2}\setcounter{totalnumber}{6}
\makeatletter\setlength{\@fptop}{0pt}\setlength{\@dblfptop}{0pt}\setlength{\@dblfpsep}{14pt plus 1fil}\setlength{\@fpsep}{12pt plus 1fil}\makeatother

\section*{Appendix}

\noindent Table~\ref{tab:map} maps each claim of the main text to the pre-registered tests that bear on it (Appendix~\ref{app:prereg}) and to the exploratory analyses, with the appendix where each is reported. Appendix tables use short condition names (rule = enforced rule, policy or stated = stated rule, cost or stated cost = call cost, think = reasoning mode), short regime names (pers.\ = persistent, rec$c$ = recover\_after\_$c$, late = late\_onset\_from\_3) and the test IDs of the pre-registration.

\begin{table*}[tp]
\centering\footnotesize\setlength{\tabcolsep}{4pt}\renewcommand{\arraystretch}{1.15}
\begin{tabular}{@{}>{\raggedright\arraybackslash}p{3.8cm} >{\raggedright\arraybackslash}p{6.8cm} >{\raggedright\arraybackslash}p{2.9cm} >{\raggedright\arraybackslash}p{1.6cm}@{}}
\toprule
Claim (main text) & Pre-registered tests (outcome) & Exploratory & Appendix \\
\midrule
Agents judge a failing source's results useless (\S\ref{sec:layers}) & 16b: stated and side-channel judgments agree (pass) & judgment rates; specificity on real results; accuracy of stated judgments & \ref{app:annot}, \ref{app:spec}, \ref{app:add16} \\
Unaided agents rarely answer after five useless judgments (\S\ref{sec:layers}) & J1 (reported); F5, fresh300 (pass 3/3); T2, stated judgments (pass 2/2); R3, Search-R1 (fail) & whether stopping would have paid & \ref{app:add12}, \ref{app:add16}, \ref{app:cap} \\
Without enforcement, stopping does not follow the useless run (\S\S\ref{sec:layers}--\ref{sec:drivers}) & J2: none, permit, budget (pass 12/12), stated rule (fail for \llama, \qthirty); C, call cost (pass 3/4); PP1, stated rule (partial for \llama only); H1 and H3 of A20, \haiku\ (partial; not on the evidence); F1--F2, fresh300 (pass 8/8); T1 of A18, stated judgments (pass 4/4) & hazard model; decide & \ref{app:add12}, \ref{app:hazard}, \ref{app:add16}, \ref{app:add19} \\
A stated budget moves stopping to the deadline (\S\ref{sec:drivers}) & H9, H10, F-H9, F-H10, D1, D3, D4, Rp3 (pass); L-H9 (fail) and A9 (mixed), \qthirty & & \ref{app:tables}, \ref{app:rep} \\
The enforced rule raises success (\S\ref{sec:integration}) & H1, F-H1, L-H1, Rp1, F4, R2, S-H2 (pass); S-H1, \sonnet\ (fail, $p=.051$) & & \ref{app:tables}, \ref{app:sonnet}, \ref{app:add16}, \ref{app:rep} \\
\quad without loss where the source recovers & H3 (14/16; 18/20 with \qthirty's four under L-H1), F-H3 (12/12); 13c, recovery after four or five failures (mixed) & & \ref{app:tables}, \ref{app:add13} \\
\quad because it acts on an accurate signal & H2, F-H2 (pass) & lexical detector & \ref{app:tables} \\
\quad with a stopping point that does not follow the budget & D2 (fail for \qtwo); A9x, \qthirty\ (pass) & & \ref{app:tables} \\
Deadline and integration combine (\S\ref{sec:integration}) & C1--C4, L-C1, L-C3 (pass); Rp4 (5/6) & & \ref{app:tables}, \ref{app:rep} \\
How good the stopping point is (\S\ref{sec:integration}) & D3, C4: tool calls (pass) & best simple policy; cost & \ref{app:cap} \\
\quad with no judgment call, read from the agent's reasoning & E1--E3 of A19 (pass on the 3 applicable models; \qtwo\ not applicable), E4 (2/3) & & \ref{app:add19} \\
The pattern replicates (\S\ref{sec:fever}) & fresh300: F1--F5 (pass); FEVER: F-H1--F-H10 (pass); second and third runs: Rp1--Rp4, 13d (Rp2 fails for \qtwo; Rp4 5/6 in the second run) & plausible and answerless sources & \ref{app:add16}, \ref{app:tables}, \ref{app:rep}, \ref{app:add12}, \ref{app:add13} \\
Larger, reasoning and search-trained agents (\S\ref{sec:scale}) & L-H1, L-H7, L-C1, L-C3 (pass), L-H9 (fail); T1 of A15, reasoning mode (pass); V, Search-R1 judgments (fail), R2 (pass), R1 (formally pass, not interpreted), R3 (fail) & \sonnet\ design-label contrast & \ref{app:tables}, \ref{app:sonnet}, \ref{app:add16} \\
\bottomrule
\end{tabular}
\caption{Where each claim of the main text is tested. Test IDs and decision rules as in Tables~\ref{tab:prereg1} and~\ref{tab:prereg}; exploratory analyses were not pre-registered.}
\label{tab:map}
\end{table*}

\section{Pre-registration, amendments and deviations}\label{app:prereg}

\paragraph{Protocol.} The pre-registration was written before any held-out run (test300 had seen only a three-question pipeline check). All thresholds were set on the 100 development questions. $k=5$ maximized development mean6 for \qthree and \llama (0.405, 0.268), and \qtwo's development sweep also peaked at $k=5$ (0.192). The \qtwo result was recorded in a results file before its test runs but was not entered in the document as the protocol required, which we count as a deviation. \haiku's $k=5$ came from its own pilot. Differences are paired by question, with 95\% intervals from 2,000 bootstrap resamples (clustered by question for \ii); tests across models are Holm-corrected; non-inferiority margins are $-0.05$. The stop rule (withdraw the intervention claim if H1 failed for two of three 7--8B models) did not trigger.

\paragraph{Amendments.} Each amendment was frozen before the runs it governs; A3's H9--H10 were added after the prompts' development runs and A9 after L-H9 had failed. A fair-coin random control almost never yields five ``useless'' in a row, so A1 replaced it by one matched to each model's development ``useless'' rate. A1 also set the lexical threshold by balanced accuracy (0.25) instead of accuracy (0.6). A2 made unparseable judgments reset the rule's counter (3.4\% of \haiku's; none for local models). A6 widened C1's margin from $-0.02$ to $-0.05$ at freezing.

\paragraph{Registration and multiplicity.} The pre-registration and its amendments were kept as one append-only document, and each amendment was frozen before the runs it governs. The document was not deposited with a third-party registry. Holm correction is applied across models within each hypothesis, not across the many hypotheses in Tables~\ref{tab:prereg1} and~\ref{tab:prereg}; we rest the central claims on effects that replicate across models, two further runs and the fresh questions rather than on any single test.

\begin{table*}[tp]
\centering\scriptsize
\setlength{\tabcolsep}{3pt}
\begin{tabular}{@{}>{\raggedright\arraybackslash}p{0.75cm}>{\raggedright\arraybackslash}p{0.75cm}>{\raggedright\arraybackslash}p{7.5cm}>{\raggedright\arraybackslash}p{6.95cm}@{}}
\toprule
Doc & ID & Prediction; models; decision rule & Outcome \\
\midrule
Orig. & H1 & Rule $>$ none (mean6); 7--8B, \haiku; one-sided, Holm & pass 4/4: $+.045$ / $+.064$ / $+.026$; \haiku $+.097$ \\
 & H2 & Rule $>$ rule fed a rate-matched random signal (A1); 7--8B; one-sided, Holm & pass 3/3: $+.011$ ($p{=}.0285$) / $+.029$ / $+.021$ \\
& H3 & Rule $\ge$ none $-0.05$ on rec1, rec2, rec3, clean; 7--8B, \haiku; 95\% LB & 14/16; \textbf{fail} \qthree rec1 (LB $-.057$), rec3 (LB $-.060$) \\
& H4 & Decide $<$ rule (mean6); 7--8B; one-sided & rule $-$ decide $-.123$ / $+.062$ / $+.048$: \textbf{fail} for \qtwo \\
& H5$^e$ & Rule vs.\ rule fed a lexical detector; 7--8B & $+.012$ / $-.001$ / $-.011$ \\
& H6$^e$ & Thinking vs.\ rule, incl.\ rec3; \qthree & $+.003$ [$-.032$, $+.039$]; rec3: .437 vs.\ rule .470 \\
\addlinespace[2pt]
A3 & H7 & Rule $>$ permit (mean6); 7--8B; one-sided, Holm & $+.023$ ($p{=}.055$) / $+.059$ / $+.051$: \textbf{fail} for \qtwo \\
 & H8 & Rule vs.\ budget (mean6); 7--8B; two-sided, no direction & $-.091$ [$-.124$, $-.060$] / $+.016$ / $+.009$ \\
& H9 & \ii upper bound $< +0.10$ for permit and for budget; 7--8B & pass 6/6 (largest bound $+.064$) \\
& H10 & Budget: over half of persistent answers at action 8; 7--8B & pass 3/3: .749 / .791 / .506 \\
\addlinespace[2pt]
A4 & F-H1 & As H1 on FEVER; 7--8B & pass 3/3: $+.129$ / $+.201$ / $+.142$ \\
 & F-H2 & As H2 on FEVER; 7--8B & pass 3/3: $+.044$ / $+.123$ / $+.096$ \\
& F-H3 & As H3 on FEVER; 7--8B & pass 12/12 \\
& F-H9 & As H9 on FEVER; 7--8B & pass 6/6 (largest bound $+.064$) \\
& F-H10 & As H10 on FEVER; 7--8B & pass 3/3: .77 / .96 / .69 \\
\addlinespace[2pt]
A5 & D1 & 16 actions, budget: median answering action on persistent $\ge 12$; 7--8B & pass 3/3: 16 / 16 / 16 \\
 & D2 & 16 actions, rule: median answering action on persistent stays 6; 7--8B & 2 / 6 / 6: \textbf{fail} for \qtwo (6 on its 107 fired questions) \\
& D3 & 16 actions: budget makes $\ge 4$ more tool calls than rule on persistent; 7--8B & pass 3/3: $+10.61$ / $+9.23$ / $+6.53$ \\
& D4 & 16 actions, budget: \ii upper bound $< +0.10$ (last action excluded); 7--8B & pass 3/3: UB $-.012$ / $+.012$ / $-.027$ (index $-.027$ / $+.003$ / $-.052$) \\
\addlinespace[2pt]
A6 & C1 & Combo $\ge$ rule $-0.05$ and $\ge$ budget $-0.05$, HotpotQA and FEVER; 7--8B & pass 12/12 (lowest LB $-.023$) \\
 & C2 & Combo $>$ rule; \qtwo; one-sided, each task & pass: $+.138$ [$+.104$, $+.173$]; FEVER $+.088$ [$+.059$, $+.118$] \\
& C3 & Combo: \ii lower bound $> +0.10$ (HotpotQA, FEVER, 16 actions); 7--8B & pass 9/9 (lowest bound $+.391$) \\
& C4 & 16 actions: combo $\ge 4$ fewer tool calls than budget, median answer $\le 7$; 7--8B & pass 3/3: 8.66 / 9.41 / 7.62 fewer; median 6 \\
\addlinespace[2pt]
A7 & S-H1 & Rule $>$ none (mean6); \sonnet, first 100 questions; one-sided & \textbf{fail}: $+.022$ [$-.003$, $+.052$], $p{=}.051$ \\
 & S-H2 & Rule $>$ none on persistent; \sonnet; one-sided & pass: $+.11$ [$+.05$, $+.18$] \\
\addlinespace[2pt]
A8 & L-H1 & As H1; \qthirty & pass: $+.063$ [$+.045$, $+.080$]; rec1--3, clean: LB $\ge -.043^\ddagger$ \\
 & L-H7 & As H7; \qthirty & pass: $+.055$ [$+.026$, $+.084$] \\
rev. & L-H9 & As H9; \qthirty & \textbf{fail} 2/2: $+.172$ [$+.108$, $+.234$], $+.317$ [$+.277$, $+.358$] \\
 & L-C1 & As C1 (HotpotQA); \qthirty & pass 2/2: LB $-.022$ (vs.\ rule), $-.019$ (vs.\ budget) \\
& L-C3 & As C3 (HotpotQA); \qthirty & pass: $+.525$ [$+.497$, $+.554$] \\
\addlinespace[2pt]
A9 & A9 & 16 actions, budget: evidence-driven (median $\le 8$, last-action share $< .2$, \ii LB $> +.10$), deadline-driven (median $\ge 12$ or share $\ge .5$), else mixed; \qthirty & \textbf{mixed}: median 7, share .01, but \ii $+.103$ [$+.064$, $+.144$] \\
 & A9x & 16 actions, rule and combo: median $\le 7$, tool calls within 1 of 8 actions; \qthirty & pass 2/2: median 6, 6; calls 4.99$\to$4.97, 4.19$\to$4.28 \\
\bottomrule
\end{tabular}
\caption{Every pre-registered test, part 1 (original pre-registration and amendments A3--A9); notation as in Table~\ref{tab:prereg}.}
\label{tab:prereg1}
\end{table*}

\begin{table*}[tp]
\centering\scriptsize
\setlength{\tabcolsep}{3pt}
\begin{tabular}{@{}>{\raggedright\arraybackslash}p{0.75cm}>{\raggedright\arraybackslash}p{0.75cm}>{\raggedright\arraybackslash}p{7.5cm}>{\raggedright\arraybackslash}p{6.95cm}@{}}
\toprule
Doc & ID & Prediction; models; decision rule & Outcome \\
\midrule
A10 & Rp1 & Replicate (new seeds, new draw of failed observations): rule $>$ none (mean6); 7--8B; one-sided, Holm & pass 3/3: $+.048$ / $+.061$ / $+.042$ \\
 & Rp2 & Replicate: rule \ii\ LB $> +.30$; none \ii\ $\le +.05$; 7--8B & rule \ii\ .19 [.14, .24] / .46 / .46: \textbf{fail} for \qtwo; none $-.31$ / $.00$ / $.00$: pass \\
& Rp3 & Replicate: budget arm's median answering action on persistent $= 8$ and \ii\ $\le +.10$; 7--8B & pass 3/3: median 8 / 8 / 8; \ii\ $-.01$ / $+.04$ / $-.02$ \\
& Rp4 & Replicate: combo $\ge$ rule $-0.05$ and $\ge$ budget $-0.05$ (95\% LB); 7--8B & 5/6: vs.\ rule $+.108$ / $+.007$ / $-.024$ [LB $-.054$], \textbf{fail} for \qthree\ (repaired: pass); vs.\ budget LB $+.033$ / $+.008$ / $+.003$ \\
\addlinespace[2pt]
A11 & PP1 & Stopping rule stated in the prompt: executed on the evidence if $\Delta$ LB $>0$, persistent median $\le 7$ and non-inferior to the rule; not executed if $\Delta$ CI includes or lies below 0 or persistent answer rate $< .5$; else partial; 7--8B, \qthirty & not executed / \textbf{partial} / not executed; \qthirty\ not executed (design-label $\Delta$ $-.05$ / $+.07$ / $-.10$; $-.10$) \\
 & PP2 & Stated rule vs.\ external rule (non-inferiority, margin $-0.05$) and vs.\ permit (mean6) & \textbf{fail} 4/4 vs.\ rule: $-.031$ / $-.061$ / $-.034$; \qthirty\ $-.047$ (LB $-.056$ / $-.089$ / $-.061$ / $-.073$; all upper bounds $<0$; repaired: \qthree\ $-.003$ [$-.026$, $+.021$], others unchanged in sign); vs.\ permit $-.007$ / $-.002$ / $+.016$; $+.008$ \\
\addlinespace[2pt]
A12 & J1 & Same-trajectory dissociation (unaided; own replayed judgments); 7--8B, \qthirty & reported: answers on 3/122, 1/282, 1/292; 19/284 \\
 & J2 & Own-judgment $\Delta$: CI includes or lies below 0 for none, permit, budget, policy, cost; LB $>0$ for rule & rule pass 4/4 ($+.33$ to $+.37$); none, permit, budget pass 12/12; \textbf{fail}: policy \llama\ ($+.09$), \qthirty\ ($+.08$); cost \qthirty\ ($+.09$) \\
& C & Stated cost: $\Delta$ CI includes or lies below 0; combo no longer called cost-optimal if cost arm's cost-adjusted success LB $> -0.02$ vs.\ combo & $\Delta$: pass 3/4, \textbf{fail} \qthirty; vs.\ combo $-.095$ / $-.039$ / $-.041$; \qthirty\ $-.003$ [$-.027$, $+.020$] (LB $< -.02$) \\
\addlinespace[2pt]
A13 & S1--S3 & Two sources: switched by action 4 $\ge .5$ $\Rightarrow$ ``switches''; else ``does not abandon'' if own-judgment $\Delta$ for leaving includes or lies below 0; 7--8B, \qthirty & \qtwo, \llama: does not abandon (.49, .17; $\Delta$ $-.54$, $-.06$); \qthree, \qthirty: switches (.73, .95) \\
 & S4 & Switch rule $-$ none, mean over six two-source regimes & \llama\ $+.054$, \qthree\ $+.016$ (both LB $>0$); \qtwo\ $+.011$, \qthirty\ $+.007$ (n.s.) \\
& 13c & Long recovery: rule $<$ none in rec5; rule $\ge$ none $-0.05$ in rec4 & rec5: holds for \qthree\ ($-.213$), \qthirty\ ($-.137$), \textbf{fails} for \qtwo, \llama; rec4: 3/4, \textbf{fail} \qthree\ (LB $-.050$) \\
& 13d & Third run: Rp1--Rp4 & Rp1 3/3; Rp2 \textbf{fail} \qtwo\ (as before); Rp3 3/3; Rp4 6/6 \\
\addlinespace[2pt]
A14 & & Both exits named: backup share of first exits $>.6$ / $<.4$; 7--8B, \qthirty & prefers evidence: \qtwo\ .93, \llama\ .93, \qthirty\ .90; no clear preference: \qthree\ .52 \\
\addlinespace[2pt]
A15 & T1 & \qthirty\ thinking mode, four regimes: own-judgment $\Delta$ CI includes or lies below 0 for none and budget; else the claim is narrowed & pass 2/2: none $+.01$ [$-.18$, $+.17$]; budget $+.25$ [$-.01$, $+.50$] (few late decisions) \\
\bottomrule
\end{tabular}
\caption{Every pre-registered test (part 2, amendments A10--A20; part 1 in Table~\ref{tab:prereg1}), by the document that fixed it (IDs and condition names as registered: policy = stated rule, rule or external rule = enforced rule, cost = call cost; Orig.: original pre-registration; A3--A20: amendments; A1 and A2 are described in the text). Test IDs are those of each document; where a document names none (A12's C, A15's T1) or its names would clash (A17's S1--S3 and validity check), we label them here (R1--R3, V). Paired differences in mean6 unless stated, as \qtwo\,/\,\llama\,/\,\qthree where three appear; brackets: 95\% intervals; LB: lower bound; 7--8B: \qtwo, \llama, \qthree; rec$c$: recover\_after\_$c$. Raw analysis. $^e$Exploratory by design. $^\ddagger$A8 adopted the local models' criteria without naming H3; the main text counts these four tests together with H3. A9's predictions are unnamed in the pre-registration.}
\label{tab:prereg}
\end{table*}

\begin{table*}[tp]
\centering\scriptsize
\setlength{\tabcolsep}{3pt}
\begin{tabular}{@{}>{\raggedright\arraybackslash}p{0.75cm}>{\raggedright\arraybackslash}p{0.75cm}>{\raggedright\arraybackslash}p{7.5cm}>{\raggedright\arraybackslash}p{6.95cm}@{}}
\toprule
Doc & ID & Prediction; models; decision rule & Outcome \\
\midrule
A16 & F1--F2 & fresh300: own-judgment $\Delta$ UB $< +.06$ for none and budget; 7--8B (\qthirty\ if time allows) & pass 8/8: none $-.09$ / $-.03$ / $-.05$, \qthirty\ $-.06$; budget $-.17$ / $-.14$ / $-.19$, \qthirty\ $-.05$ (UB $+.01$; others $<0$) \\
 & F3 & fresh300: rule $\Delta$ LB $> +.20$ & pass 4/4: $+.31$ [LB $+.22$] / $+.35$ [$+.33$] / $+.34$ [$+.31$]; \qthirty\ $+.30$ [$+.26$] \\
& F4 & fresh300: rule $>$ none (mean6), one-sided, Holm & pass 4/4: $+.030$ / $+.061$ / $+.036$; \qthirty\ $+.057$ (all $p<.001$) \\
& F5 & fresh300: unaided agent answers on $< 10\%$ of questions after five own useless judgments (persistent) & pass 3/3: 3/125, 2/283, 0/289 \\
& 16b & Stated vs.\ side-channel judgments: annotator binary $\kappa \ge .80$ on the adjudicated labels to proceed; agreement $\ge .85$ expected & $\kappa$ .874 [.798, .939]; agreement .87 / .94 / .94; \qthirty\ .96 \\
\addlinespace[2pt]
A17 & V & Search-R1: judgment validity, USELESS after failed $\ge .80$ and after real $\le .50$, else $\Delta$ only descriptive & \textbf{fail}: 1.00 and .89 \\
 & R1 & Search-R1: own-judgment $\Delta$ UB $< +.06$ (none) & formally pass ($-.26$ [$-.58$, $-.06$]); not interpreted (V) \\
& R2 & Search-R1: rule $>$ none (mean6), one-sided & pass: $+.024$ [$+.009$, $+.041$] \\
& R3 & Search-R1: unaided answers on $< 10\%$ after five own useless judgments (persistent) & \textbf{fail}: 29/165 (.18) \\
\addlinespace[2pt]
A18 & T1 & $\Delta$ from stated judgments (unaided, test300): UB $< +.06$ & pass 4/4: $-.08$ / $-.01$ / $-.04$; \qthirty\ $-.02$ (UB $+.002$) \\
 & T2 & Stated judgments: answers on $< 10\%$ after five consecutive stated useless judgments (persistent); descriptive below 20 questions and for \qthirty & pass 2/2: \llama\ 1/266, \qthree\ 1/287; \qtwo\ 0/15, \qthirty\ 16/252 (descriptive) \\
A19 & E1 & Stated-judgment rule (keyword reader on the agent's own thought, no judgment call) $>$ none (mean6); one-sided, Holm & pass 3/3 applicable: $+.064$ / $+.036$ / $+.064$ (\llama, \qthree, \qthirty); \qtwo\ $+.018$, not applicable (fires on 0.05 of persistent questions) \\
 & E2 & Stated-judgment rule $\ge$ enforced rule $-0.05$ (mean6, 95\% LB) & pass 3/3 applicable (lowest LB $-.017$); \qtwo\ LB $-.041$ \\
& E3 & Utility at $\lambda=0.05$, all calls counted: stated-judgment rule $>$ enforced rule; one-sided, Holm & pass 3/3 applicable: $+.236$ / $+.196$ / $+.190$; \qtwo\ $+.086$ \\
& E4 & Utility at $\lambda=0.05$: stated-judgment combination $>$ budget; one-sided, Holm & 2/3 applicable: $+.045$ / $+.022$; \textbf{fail} \qthirty\ ($.000$); \qtwo\ $+.027$ \\
A20 & H1 & \haiku, rule stated in the prompt (200 questions): executed / partial / not executed, by the rule of PP1 & \textbf{partial}: design-label contrast $+.052$ [$+.021, +.085$], median action 7, vs.\ enforced rule $-.050$ [LB $-.086$] \\
 & H2 & \haiku, stated rule vs.\ none and vs.\ enforced rule (mean6), reported & $+.043$ / $-.050$ \\
& H3 & \haiku, call cost stated: design-label contrast CI includes or lies below 0; utility at $\lambda=0.05$ reported & pass: $+.013$ [$-.023, +.048$]; mean6 vs.\ enforced rule $+.053$; utility vs.\ none $+.154$, vs.\ enforced rule $+.213$ \\
\bottomrule
\end{tabular}
\addtocounter{table}{-1}
\caption[]{(continued) Amendments A16--A20.}
\end{table*}

\paragraph{Deviation 1 (during the original runs).} One trajectory of \qtwo's unaided arm exceeded the 8,192-token context, so that arm was run with 16,384 tokens and the same greedy decoding; all other \qtwo arms stayed within 8,192 tokens, so the limit could not have changed them. This had no effect.

\paragraph{Deviation 2 (found after the runs).} Some answers copied the template \texttt{finish[answer]}; they were repaired uniformly (Appendix~\ref{app:repair}).

\paragraph{Deviation 3 (during A5).} \qtwo's 16-action budget arm exceeded 16,384 tokens in recover\_after\_2; only that regime was rerun, with 32,768 tokens and greedy decoding. This had no effect.

\paragraph{Deviation 4 (executing A15).} A15's third arm (stated cost) was not run; A15's predictions concern only the unaided and budget arms, which ran as planned.

\paragraph{Deviation 5 (executing A16).} On fresh300, as in deviation 1, some trajectories exceeded the 8,192-token context, so the \qthree\ arms, their side-channel replays and the remaining fresh300 runs used 16,384 tokens (\llama's none and budget arms had completed at 8,192). The limit affects only trajectories that would otherwise exceed it.

\paragraph{Deviation 6 (executing A15).} The \qthirty\ thinking runs allowed up to 8,192 new tokens per step instead of the registered 2,048; 25 of 7,141 generations exceeded 2,048.

\paragraph{Exploratory analyses.} Not pre-registered: \ii\ in the first runs for the unaided, decide, thinking, random, lexical, call-cost and rule arms (it was pre-registered for the permission, budget, stated-rule and combination arms, as a descriptive quantity for unaided \qthirty and both \sonnet arms, and for the rule and unaided arms of the second run); the combination's superiority; FEVER comparisons of the rule with lexical, decide and prompt arms; thinking's token cost; beliefs on test300; \sonnet's non-inferiority; and a noise check. Rerunning unaided \qthree and \llama with identical settings reproduced them (mean6 difference $+0.001$ each), so run-to-run variation arises when an intervention changes vLLM's batching. In \qthree's failed H3 tests, questions where the rule fired account for a net $-2$ and $-5$ correct answers of 300 (rec1, rec3), questions it never touched for $-6$ and $-4$.

\section{Setup details}\label{app:setup}

\subsection{Prompts and signals}\label{app:prompts}

\paragraph{Agent.} All HotpotQA runs use the system prompt below, except Search-R1, which keeps the prompt it was trained with (Appendix~\ref{app:add16}); the two-source runs add the backup tool (Appendix~\ref{app:add13}). FEVER replaces the task description, the Thought line and the answer line (``Submit your verdict. The answer must be exactly one of: SUPPORTS, REFUTES''), opens with ``Verify this claim:'', and says ``claim''/``verdict'' in the judgment, answer-now and action prompts; the budget and cost statements are unchanged.

\begin{quote}\footnotesize\ttfamily
You are a question-answering agent. You answer questions by interacting with a Wikipedia-like knowledge base.\\[2pt]
Available actions:\\
- search[query]: Search for a Wikipedia article. Returns the article text.\\
- lookup[keyword]: Find the next sentence containing the keyword in the current article.\\
- finish[answer]: Submit your final answer. The answer should be short (a few words).\\[2pt]
Format your response as:\\
Thought: <your step-by-step reasoning>\\
Action: <action\_name>\\
Action Input: <input>\\[2pt]
Produce exactly one Thought and one Action per turn. End your response after Action Input and wait for the tool observation. Do not simulate tool results or future actions.
\end{quote}

\paragraph{Prompt arms.} \emph{permit} appends to the system prompt:
\begin{quote}\footnotesize\ttfamily
If the search results do not give you the information you need, you may also answer from your own knowledge: use finish[answer] with your best guess.
\end{quote}
\emph{budget} appends the same sentence and
\begin{quote}\footnotesize\ttfamily
You can take at most 8 actions for each question, including finish. If you have not used finish[answer] by then, the question is scored as wrong.
\end{quote}
(16 in the doubled-budget runs), and adds \texttt{[Actions remaining: $n$]} to the question and to every observation.
The \emph{stated rule} appends to the permit prompt
\begin{quote}\footnotesize\ttfamily
Stopping policy: if five consecutive search or lookup results in a row are useless for answering the question, stop consulting the knowledge base and answer from your own knowledge with finish[answer].
\end{quote}
and \emph{cost} appends to the budget prompt
\begin{quote}\footnotesize\ttfamily
Scoring: a correct final answer earns 1 point; each search or lookup costs 0.05 points; a wrong or missing answer earns 0. Maximize your total points.
\end{quote}

\paragraph{Side-channel judgment.} Before each action after the first, the history is copied, the copy receives
\begin{quote}\footnotesize\ttfamily
Before acting: was the LAST observation useful for answering the original question? Reply with exactly one word: USEFUL or USELESS.
\end{quote}
and the reply (at most 8 tokens) is parsed and discarded; the acting history is not changed. A reply that is not an explicit USELESS resets the streak.

\paragraph{Rule.} When the streak of USELESS judgments reaches $k=5$, the system prompt receives ``For this turn only, the search and lookup actions are unavailable; the only available action is finish[answer].'' and the latest observation is followed by ``Do not make any further tool calls. Give your final answer to the original question using finish[answer].'' Otherwise the history is untouched. \emph{random} replaces the judgment by a seeded draw that says USELESS with the model's own development-set rate of USELESS judgments; \emph{lexical} says USELESS when fewer than 25\% of the question's content words occur in the first 400 characters of the observation (threshold $\tau=0.25$, chosen on development labels; balanced accuracy 0.77), or when the tool returns no result. \emph{combo} applies the budget prompt and the rule together.

\paragraph{Decide.} The agent's judgment is written into the history, followed by
\begin{quote}\footnotesize\ttfamily
Your judgment of the last observation: \{j\}. Consecutive observations judged USELESS so far: \{k\}. In one or two sentences, weigh how likely you are to succeed if you keep consulting this knowledge base against how likely you are to answer correctly right now from what you already know. Then, on a final line, write exactly: DECISION: CONTINUE or DECISION: ANSWER\_NOW.
\end{quote}
and an instruction to act consistently with the decision.

\paragraph{Belief probe (continuation turn).}
\begin{quote}\footnotesize\ttfamily
Pause. Before deciding anything, give an honest assessment of your situation. You have an unspecified number of tool interactions left with this knowledge base. Estimate two probabilities, each between 0 and 1: - p\_continue: the probability that you would end up answering the original question correctly if you kept consulting this knowledge base with the remaining interactions; - p\_answer\_now: the probability that you would answer correctly if you gave your best answer right now from what you already know. Reply with a single JSON object and nothing else: \{"p\_continue": number, "p\_answer\_now": number, "prefer": one of "continue", "answer\_now", "unsure", "reason": one short sentence\}
\end{quote}

\subsection{Annotation of stated judgments}\label{app:annot}
On \qthree's development trajectories, two label streams were annotated from a written rubric: the evidence type of each observation (S1: useful new, related but not answering, redundant, off topic, unclear) and the judgment the agent states in its next thought (S2: says useless, partly useful, useful, no explicit judgment, unclear). The reference set below used rubric v0.3.1; v0.3.2, released with the code, adds the rulings from its adjudication. Records were shown blinded, without regime, gold answer or outcome.

\paragraph{Full set.} Two LLM annotators of different families, \sonnet (primary) and GPT-5.6-luna, labeled all 1,844 observations and 1,675 stated judgments independently through their official APIs. On the 1,792 and 1,672 items both labeled validly they agree at five-class $\kappa$ 0.617 (S1) and 0.844 (S2), binary 0.927 (useful new or not) and 0.903 (says useless or not); items of the reference set below take their adjudicated label. Where they agree, their label is used. The first author adjudicated all 59 disagreements that change a consecutive-useless count. For the 328 that do not (mostly off topic versus related), the primary annotator's label is used. The first author also adjudicated a random 40 of these and found the primary label correct on 35 (0.875, Wilson 95\% [0.74, 0.95]); no count changes.

\paragraph{Reference set.} Two further LLM annotators, Claude Fable 5.1 and GPT-6-sol, independently labeled a stratified sample of 360 observations and 327 stated judgments through their official APIs, without access to each other's labels. They agree at $\kappa$ 0.875 and 0.868 (binary 0.972 and 0.898), and the first author adjudicated their disagreements. No human annotators other than the first author were involved.

\paragraph{Accuracy of stated judgments.} Against these labels, the judgments \qthree states in its own reasoning call useless evidence useless 93.9\% of the time ($n=1{,}463$) and treat useful or partly useful evidence as such 98.1\% of the time ($n=212$).

\paragraph{Stated versus side-channel judgments.} For amendment 16b a local \qthirty annotator (non-thinking, temperature 0, the same rubric) labeled the 327 reference stated judgments at five-class $\kappa$ 0.842 and binary 0.874 before being applied to four models' test trajectories (Appendix~\ref{app:add16}).

\subsection{Figure 1 selection rule}\label{app:fig1}
Held-out questions were taken in the order of the released question manifest and the first one was selected where (1) \haiku without intervention exhausts its budget in the persistent regime, (2) on the same question the rule arm's side-channel judgments are USELESS five times in a row and the rule fires, and (3) the rule arm's answer is correct. No other candidate was inspected.

\subsection{Claude Sonnet 5}\label{app:sonnet}

\paragraph{Setup.} Following A7, \sonnet ran through the official API without thinking (it accepts no sampling parameters, so decoding was the service default) on the first 100 test300 questions, in the six mean6 regimes, with the unaided and rule arms. The rule run's per-step side-channel judgments were not saved for five of its six regimes, so the pre-registered share of ``useless'' judgments and its own-judgment measures cannot be computed. Rule firings are recovered from the trajectories. A firing requires five consecutive USELESS judgments and appends the strict forced-answer suffix (``\ldots the only available action is finish[answer].'') to the system message. The rule fired on 65 of 100 persistent questions.

\paragraph{Pre-registration and results.} A7 fixed S-H1 (rule $-$ none, mean6) and S-H2 (persistent), both one-sided; descriptive quantities (the ``useless'' share, unaided answering after five ``useless'' judgments, both arms' \ii); and two rules: report no significant gain if S-H1 failed, and partial self-stopping, judged by \ii, if the unaided agent answered on most persistent questions. S-H2 passed ($+0.11$ [$+0.05$, $+0.18$]); S-H1 failed ($+0.022$ [$-0.003$, $+0.052$], $p=0.051$). The unaided agent answered on 69 of 100 persistent questions, with \ii $+0.18$ [$+0.11$, $+0.26$] (rule: $+0.49$); on the 65 questions where the rule fired, the unaided agent exhausted its budget on 29 (45\%; \haiku: 282 of 284).

\paragraph{Non-inferiority (not pre-registered).} Rule $-$ none was $+0.00$ [$-0.06$, $+0.06$], $+0.02$ [$-0.04$, $+0.08$] and $-0.03$ [$-0.09$, $+0.03$] on recover\_after\_1--3 and $-0.01$ [$-0.08$, $+0.06$] on clean; only recover\_after\_2 clears $-0.05$. Non-inferiority therefore cannot be shown in three of four regimes. This reflects sample size and run-to-run variation rather than harm. Premature answers in recover\_after\_1--3 occur before the rule can fire, yet they also differed between arms (0.05, 0.12, 0.18 with the rule; 0.00, 0.07, 0.17 without).

\section{Measures}\label{app:measures}

\subsection{Time-matched contrast}\label{app:add12}
Amendment 12 was frozen before any of its runs. For the unaided, permission, budget, stated-rule and stated-cost arms, the side-channel question of the rule arm was asked again at every decision point of the recorded trajectories (the recorded history verbatim, at most 8 generated tokens, nothing written back); the rule and combination arms use the judgments recorded in their own runs. In the same trajectories, once its own judgments have called five consecutive results useless, the unaided agent answers on 3 of 122 questions (\qtwo), 1 of 282 (\llama), 1 of 292 (\qthree) and 19 of 284 (\qthirty).

\paragraph{Definition.} After $t$ results, let $s$ be the current run of consecutive results the agent judged useless (any other reply resets it). At each decision with $3 \le t \le 6$ (actions 4--7), group (a) has $s=t$, every result judged useless, and the reference group (b) has $1 \le s \le t-2$: the latest result judged useless and the most recent result not judged useless lying between the second and the second-to-last. Excluding $s=t-1$ keeps (b) from differing from (a) only in the first result. $\Delta$ is the Mantel--Haenszel risk difference in answering between (a) and (b), stratified by $t$; intervals come from 2,000 question-level bootstrap resamples.

Table~\ref{tab:deltaown} gives the time-matched contrast on the agent's own judgments (the measure in Figure~\ref{fig:delta}), the same contrast stratified by question and action so that every comparison is within a question, and the exploratory version on design labels, with 95\% question-cluster intervals. Pre-registered predictions: $\Delta$'s interval includes 0 or lies below it for the unaided and prompt arms, and lies above 0 for the rule. Both hold for the rule and for the unaided, permission and budget arms on all four models; they fail for the stated rule on \llama\ and \qthirty\ and for the stated cost on \qthirty, which the main text reports as partly evidence-driven. Within questions the stated rule stays positive for \llama\ but the two \qthirty\ effects lose significance; within regimes the stated rule's contrast for \qthirty\ grows ($+0.17$) and the stated cost's vanishes, and permission reaches $+0.11$ [$+0.00, +0.20$]. Under every stratification, the unaided and budget-cued contrasts of every model have 95\% upper bounds below $+0.06$. The stated cost (0.05 per call, with the budget) left the 7--8B models answering at the deadline (median action 8; 53--86\% of failing-source answers at the final action) and did not change cost-adjusted success relative to the budget prompt ($-0.009$ to $+0.012$); the combination's cost-adjusted success exceeds the stated cost's for the 7--8B models ($+0.039$ to $+0.095$) and ties it for \qthirty\ ($+0.003$ [$-0.020, +0.027$]). Charging each one-word judgment call of the rule and combination at the price of a tool call (second number in the last block) makes them the most expensive arms. The rule's own-judgment contrast for \haiku\ is $+0.40$ [$+0.35, +0.46$].

\paragraph{Validation on simulated policies.} Table~\ref{tab:sim} applies the three measures to simulated agents in the same five failure regimes and 8-action budget (no model calls; 300 questions, mean of 10 simulations). Corrupted observations are judged useless with probability 0.99 and real ones with probability 0.45, and a real observation judged useful ends the episode with probability 0.3; each policy adds its own stopping rule. The integration index is positive for a clock and for a hazard that rises with time, and undefined when stopping precedes four failures; both contrasts are zero in expectation for every time- or deadline-driven policy and positive for every evidence-driven policy whose threshold falls within the budget. No measure detects a threshold beyond the budget ($k=7$).

\paragraph{Non-inferiority margin.} At the $-0.02$ margin drafted before amendment 6, the combination's non-inferiority to its components holds in 12 of 14 original tests; it fails for \qthree against the rule on HotpotQA (lower bound $-0.023$) and \qthirty against the rule on HotpotQA (lower bound $-0.022$).

\begin{table*}[tp]
\centering\footnotesize\setlength{\tabcolsep}{2.5pt}
\begin{adjustbox}{max width=\linewidth}\begin{tabular}{@{}lccccccc@{}}
\toprule
Model & none & permit & budget & stated & cost & enforced & combo \\
\midrule
\multicolumn{8}{@{}l}{time-matched contrast $\Delta$ on the agent's own judgments (amendment 12)} \\
\qtwo & \begin{tabular}[c]{@{}c@{}}$-0.11$\\[-2pt]{\scriptsize[$-0.17, -0.05$]}\end{tabular} & \begin{tabular}[c]{@{}c@{}}$-0.16$\\[-2pt]{\scriptsize[$-0.31, -0.06$]}\end{tabular} & \begin{tabular}[c]{@{}c@{}}$-0.18$\\[-2pt]{\scriptsize[$-0.25, -0.11$]}\end{tabular} & \begin{tabular}[c]{@{}c@{}}$-0.31$\\[-2pt]{\scriptsize[$-0.55, -0.13$]}\end{tabular} & \begin{tabular}[c]{@{}c@{}}$-0.18$\\[-2pt]{\scriptsize[$-0.25, -0.11$]}\end{tabular} & \begin{tabular}[c]{@{}c@{}}$+0.33$\\[-2pt]{\scriptsize[$+0.25, +0.41$]}\end{tabular} & \begin{tabular}[c]{@{}c@{}}$+0.21$\\[-2pt]{\scriptsize[$+0.11, +0.29$]}\end{tabular} \\
\llama & \begin{tabular}[c]{@{}c@{}}$-0.02$\\[-2pt]{\scriptsize[$-0.03, -0.02$]}\end{tabular} & \begin{tabular}[c]{@{}c@{}}$-0.19$\\[-2pt]{\scriptsize[$-0.23, -0.16$]}\end{tabular} & \begin{tabular}[c]{@{}c@{}}$-0.10$\\[-2pt]{\scriptsize[$-0.12, -0.08$]}\end{tabular} & \begin{tabular}[c]{@{}c@{}}$+0.09$\\[-2pt]{\scriptsize[$+0.06, +0.13$]}\end{tabular} & \begin{tabular}[c]{@{}c@{}}$-0.17$\\[-2pt]{\scriptsize[$-0.19, -0.14$]}\end{tabular} & \begin{tabular}[c]{@{}c@{}}$+0.37$\\[-2pt]{\scriptsize[$+0.36, +0.39$]}\end{tabular} & \begin{tabular}[c]{@{}c@{}}$+0.27$\\[-2pt]{\scriptsize[$+0.25, +0.30$]}\end{tabular} \\
\qthree & \begin{tabular}[c]{@{}c@{}}$-0.06$\\[-2pt]{\scriptsize[$-0.10, -0.04$]}\end{tabular} & \begin{tabular}[c]{@{}c@{}}$-0.32$\\[-2pt]{\scriptsize[$-0.41, -0.24$]}\end{tabular} & \begin{tabular}[c]{@{}c@{}}$-0.20$\\[-2pt]{\scriptsize[$-0.26, -0.15$]}\end{tabular} & \begin{tabular}[c]{@{}c@{}}$-0.18$\\[-2pt]{\scriptsize[$-0.25, -0.11$]}\end{tabular} & \begin{tabular}[c]{@{}c@{}}$-0.19$\\[-2pt]{\scriptsize[$-0.24, -0.14$]}\end{tabular} & \begin{tabular}[c]{@{}c@{}}$+0.33$\\[-2pt]{\scriptsize[$+0.29, +0.37$]}\end{tabular} & \begin{tabular}[c]{@{}c@{}}$+0.20$\\[-2pt]{\scriptsize[$+0.13, +0.27$]}\end{tabular} \\
\qthirty & \begin{tabular}[c]{@{}c@{}}$-0.07$\\[-2pt]{\scriptsize[$-0.10, -0.04$]}\end{tabular} & \begin{tabular}[c]{@{}c@{}}$+0.01$\\[-2pt]{\scriptsize[$-0.08, +0.09$]}\end{tabular} & \begin{tabular}[c]{@{}c@{}}$-0.00$\\[-2pt]{\scriptsize[$-0.06, +0.05$]}\end{tabular} & \begin{tabular}[c]{@{}c@{}}$+0.08$\\[-2pt]{\scriptsize[$+0.01, +0.16$]}\end{tabular} & \begin{tabular}[c]{@{}c@{}}$+0.09$\\[-2pt]{\scriptsize[$+0.04, +0.14$]}\end{tabular} & \begin{tabular}[c]{@{}c@{}}$+0.33$\\[-2pt]{\scriptsize[$+0.29, +0.36$]}\end{tabular} & \begin{tabular}[c]{@{}c@{}}$+0.19$\\[-2pt]{\scriptsize[$+0.12, +0.25$]}\end{tabular} \\
\midrule
\multicolumn{8}{@{}l}{same, stratified by question $\times$ action (exploratory)} \\
\qtwo & \begin{tabular}[c]{@{}c@{}}$-0.17$\\[-2pt]{\scriptsize[$-0.23, -0.11$]}\end{tabular} & \begin{tabular}[c]{@{}c@{}}$-0.20$\\[-2pt]{\scriptsize[$-0.37, -0.07$]}\end{tabular} & \begin{tabular}[c]{@{}c@{}}$-0.20$\\[-2pt]{\scriptsize[$-0.28, -0.13$]}\end{tabular} & \begin{tabular}[c]{@{}c@{}}$-0.24$\\[-2pt]{\scriptsize[$-0.55, -0.02$]}\end{tabular} & \begin{tabular}[c]{@{}c@{}}$-0.23$\\[-2pt]{\scriptsize[$-0.29, -0.16$]}\end{tabular} & \begin{tabular}[c]{@{}c@{}}$+0.31$\\[-2pt]{\scriptsize[$+0.22, +0.40$]}\end{tabular} & \begin{tabular}[c]{@{}c@{}}$+0.13$\\[-2pt]{\scriptsize[$+0.02, +0.22$]}\end{tabular} \\
\llama & \begin{tabular}[c]{@{}c@{}}$-0.02$\\[-2pt]{\scriptsize[$-0.03, -0.01$]}\end{tabular} & \begin{tabular}[c]{@{}c@{}}$-0.20$\\[-2pt]{\scriptsize[$-0.24, -0.17$]}\end{tabular} & \begin{tabular}[c]{@{}c@{}}$-0.12$\\[-2pt]{\scriptsize[$-0.14, -0.10$]}\end{tabular} & \begin{tabular}[c]{@{}c@{}}$+0.13$\\[-2pt]{\scriptsize[$+0.09, +0.17$]}\end{tabular} & \begin{tabular}[c]{@{}c@{}}$-0.17$\\[-2pt]{\scriptsize[$-0.20, -0.14$]}\end{tabular} & \begin{tabular}[c]{@{}c@{}}$+0.37$\\[-2pt]{\scriptsize[$+0.34, +0.39$]}\end{tabular} & \begin{tabular}[c]{@{}c@{}}$+0.26$\\[-2pt]{\scriptsize[$+0.23, +0.29$]}\end{tabular} \\
\qthree & \begin{tabular}[c]{@{}c@{}}$-0.08$\\[-2pt]{\scriptsize[$-0.11, -0.05$]}\end{tabular} & \begin{tabular}[c]{@{}c@{}}$-0.39$\\[-2pt]{\scriptsize[$-0.48, -0.29$]}\end{tabular} & \begin{tabular}[c]{@{}c@{}}$-0.23$\\[-2pt]{\scriptsize[$-0.28, -0.18$]}\end{tabular} & \begin{tabular}[c]{@{}c@{}}$-0.20$\\[-2pt]{\scriptsize[$-0.28, -0.12$]}\end{tabular} & \begin{tabular}[c]{@{}c@{}}$-0.19$\\[-2pt]{\scriptsize[$-0.25, -0.14$]}\end{tabular} & \begin{tabular}[c]{@{}c@{}}$+0.29$\\[-2pt]{\scriptsize[$+0.24, +0.34$]}\end{tabular} & \begin{tabular}[c]{@{}c@{}}$+0.16$\\[-2pt]{\scriptsize[$+0.09, +0.24$]}\end{tabular} \\
\qthirty & \begin{tabular}[c]{@{}c@{}}$-0.07$\\[-2pt]{\scriptsize[$-0.10, -0.04$]}\end{tabular} & \begin{tabular}[c]{@{}c@{}}$-0.01$\\[-2pt]{\scriptsize[$-0.11, +0.09$]}\end{tabular} & \begin{tabular}[c]{@{}c@{}}$-0.02$\\[-2pt]{\scriptsize[$-0.08, +0.05$]}\end{tabular} & \begin{tabular}[c]{@{}c@{}}$+0.02$\\[-2pt]{\scriptsize[$-0.06, +0.10$]}\end{tabular} & \begin{tabular}[c]{@{}c@{}}$+0.07$\\[-2pt]{\scriptsize[$-0.00, +0.15$]}\end{tabular} & \begin{tabular}[c]{@{}c@{}}$+0.30$\\[-2pt]{\scriptsize[$+0.26, +0.33$]}\end{tabular} & \begin{tabular}[c]{@{}c@{}}$+0.13$\\[-2pt]{\scriptsize[$+0.05, +0.20$]}\end{tabular} \\
\midrule
\multicolumn{8}{@{}l}{same, stratified by regime $\times$ action (exploratory)} \\
\qtwo & \begin{tabular}[c]{@{}c@{}}$-0.07$\\[-2pt]{\scriptsize[$-0.13, -0.01$]}\end{tabular} & \begin{tabular}[c]{@{}c@{}}$-0.08$\\[-2pt]{\scriptsize[$-0.23, +0.03$]}\end{tabular} & \begin{tabular}[c]{@{}c@{}}$-0.14$\\[-2pt]{\scriptsize[$-0.21, -0.07$]}\end{tabular} & \begin{tabular}[c]{@{}c@{}}$-0.27$\\[-2pt]{\scriptsize[$-0.48, -0.11$]}\end{tabular} & \begin{tabular}[c]{@{}c@{}}$-0.13$\\[-2pt]{\scriptsize[$-0.19, -0.06$]}\end{tabular} & \begin{tabular}[c]{@{}c@{}}$+0.37$\\[-2pt]{\scriptsize[$+0.29, +0.44$]}\end{tabular} & \begin{tabular}[c]{@{}c@{}}$+0.23$\\[-2pt]{\scriptsize[$+0.13, +0.31$]}\end{tabular} \\
\llama & \begin{tabular}[c]{@{}c@{}}$-0.02$\\[-2pt]{\scriptsize[$-0.03, -0.01$]}\end{tabular} & \begin{tabular}[c]{@{}c@{}}$-0.05$\\[-2pt]{\scriptsize[$-0.10, -0.00$]}\end{tabular} & \begin{tabular}[c]{@{}c@{}}$-0.02$\\[-2pt]{\scriptsize[$-0.06, +0.01$]}\end{tabular} & \begin{tabular}[c]{@{}c@{}}$+0.09$\\[-2pt]{\scriptsize[$+0.03, +0.16$]}\end{tabular} & \begin{tabular}[c]{@{}c@{}}$-0.06$\\[-2pt]{\scriptsize[$-0.10, -0.01$]}\end{tabular} & \begin{tabular}[c]{@{}c@{}}$+0.36$\\[-2pt]{\scriptsize[$+0.33, +0.39$]}\end{tabular} & \begin{tabular}[c]{@{}c@{}}$+0.33$\\[-2pt]{\scriptsize[$+0.30, +0.36$]}\end{tabular} \\
\qthree & \begin{tabular}[c]{@{}c@{}}$-0.05$\\[-2pt]{\scriptsize[$-0.08, -0.02$]}\end{tabular} & \begin{tabular}[c]{@{}c@{}}$-0.20$\\[-2pt]{\scriptsize[$-0.30, -0.10$]}\end{tabular} & \begin{tabular}[c]{@{}c@{}}$-0.11$\\[-2pt]{\scriptsize[$-0.19, -0.04$]}\end{tabular} & \begin{tabular}[c]{@{}c@{}}$-0.00$\\[-2pt]{\scriptsize[$-0.10, +0.10$]}\end{tabular} & \begin{tabular}[c]{@{}c@{}}$-0.09$\\[-2pt]{\scriptsize[$-0.15, -0.02$]}\end{tabular} & \begin{tabular}[c]{@{}c@{}}$+0.37$\\[-2pt]{\scriptsize[$+0.33, +0.41$]}\end{tabular} & \begin{tabular}[c]{@{}c@{}}$+0.23$\\[-2pt]{\scriptsize[$+0.16, +0.29$]}\end{tabular} \\
\qthirty & \begin{tabular}[c]{@{}c@{}}$-0.07$\\[-2pt]{\scriptsize[$-0.11, -0.04$]}\end{tabular} & \begin{tabular}[c]{@{}c@{}}$+0.11$\\[-2pt]{\scriptsize[$+0.00, +0.20$]}\end{tabular} & \begin{tabular}[c]{@{}c@{}}$-0.07$\\[-2pt]{\scriptsize[$-0.14, +0.00$]}\end{tabular} & \begin{tabular}[c]{@{}c@{}}$+0.17$\\[-2pt]{\scriptsize[$+0.07, +0.26$]}\end{tabular} & \begin{tabular}[c]{@{}c@{}}$+0.04$\\[-2pt]{\scriptsize[$-0.05, +0.14$]}\end{tabular} & \begin{tabular}[c]{@{}c@{}}$+0.34$\\[-2pt]{\scriptsize[$+0.29, +0.37$]}\end{tabular} & \begin{tabular}[c]{@{}c@{}}$+0.18$\\[-2pt]{\scriptsize[$+0.10, +0.26$]}\end{tabular} \\
\midrule
\multicolumn{8}{@{}l}{design-label contrast (persistent $-$ late\_onset\_from\_3 at the same action; exploratory)} \\
\qtwo & \begin{tabular}[c]{@{}c@{}}$-0.03$\\[-2pt]{\scriptsize[$-0.05, -0.02$]}\end{tabular} & \begin{tabular}[c]{@{}c@{}}$-0.11$\\[-2pt]{\scriptsize[$-0.15, -0.08$]}\end{tabular} & \begin{tabular}[c]{@{}c@{}}$-0.05$\\[-2pt]{\scriptsize[$-0.07, -0.03$]}\end{tabular} & \begin{tabular}[c]{@{}c@{}}$-0.05$\\[-2pt]{\scriptsize[$-0.07, -0.03$]}\end{tabular} & \begin{tabular}[c]{@{}c@{}}$-0.06$\\[-2pt]{\scriptsize[$-0.08, -0.03$]}\end{tabular} & \begin{tabular}[c]{@{}c@{}}$+0.08$\\[-2pt]{\scriptsize[$+0.04, +0.11$]}\end{tabular} & \begin{tabular}[c]{@{}c@{}}$+0.06$\\[-2pt]{\scriptsize[$+0.02, +0.09$]}\end{tabular} \\
\llama & \begin{tabular}[c]{@{}c@{}}$-0.01$\\[-2pt]{\scriptsize[$-0.01, -0.00$]}\end{tabular} & \begin{tabular}[c]{@{}c@{}}$-0.03$\\[-2pt]{\scriptsize[$-0.06, -0.01$]}\end{tabular} & \begin{tabular}[c]{@{}c@{}}$-0.01$\\[-2pt]{\scriptsize[$-0.03, +0.01$]}\end{tabular} & \begin{tabular}[c]{@{}c@{}}$+0.07$\\[-2pt]{\scriptsize[$+0.04, +0.11$]}\end{tabular} & \begin{tabular}[c]{@{}c@{}}$-0.03$\\[-2pt]{\scriptsize[$-0.05, -0.02$]}\end{tabular} & \begin{tabular}[c]{@{}c@{}}$+0.19$\\[-2pt]{\scriptsize[$+0.16, +0.21$]}\end{tabular} & \begin{tabular}[c]{@{}c@{}}$+0.19$\\[-2pt]{\scriptsize[$+0.16, +0.22$]}\end{tabular} \\
\qthree & \begin{tabular}[c]{@{}c@{}}$-0.02$\\[-2pt]{\scriptsize[$-0.03, -0.01$]}\end{tabular} & \begin{tabular}[c]{@{}c@{}}$-0.15$\\[-2pt]{\scriptsize[$-0.26, -0.07$]}\end{tabular} & \begin{tabular}[c]{@{}c@{}}$-0.11$\\[-2pt]{\scriptsize[$-0.15, -0.07$]}\end{tabular} & \begin{tabular}[c]{@{}c@{}}$-0.10$\\[-2pt]{\scriptsize[$-0.16, -0.05$]}\end{tabular} & \begin{tabular}[c]{@{}c@{}}$-0.08$\\[-2pt]{\scriptsize[$-0.12, -0.03$]}\end{tabular} & \begin{tabular}[c]{@{}c@{}}$+0.16$\\[-2pt]{\scriptsize[$+0.12, +0.20$]}\end{tabular} & \begin{tabular}[c]{@{}c@{}}$+0.05$\\[-2pt]{\scriptsize[$-0.00, +0.11$]}\end{tabular} \\
\qthirty & \begin{tabular}[c]{@{}c@{}}$-0.02$\\[-2pt]{\scriptsize[$-0.04, -0.00$]}\end{tabular} & \begin{tabular}[c]{@{}c@{}}$-0.06$\\[-2pt]{\scriptsize[$-0.15, +0.02$]}\end{tabular} & \begin{tabular}[c]{@{}c@{}}$-0.03$\\[-2pt]{\scriptsize[$-0.09, +0.02$]}\end{tabular} & \begin{tabular}[c]{@{}c@{}}$-0.10$\\[-2pt]{\scriptsize[$-0.20, -0.01$]}\end{tabular} & \begin{tabular}[c]{@{}c@{}}$+0.05$\\[-2pt]{\scriptsize[$-0.01, +0.11$]}\end{tabular} & \begin{tabular}[c]{@{}c@{}}$+0.17$\\[-2pt]{\scriptsize[$+0.13, +0.21$]}\end{tabular} & \begin{tabular}[c]{@{}c@{}}$+0.05$\\[-2pt]{\scriptsize[$-0.02, +0.11$]}\end{tabular} \\
\midrule
\multicolumn{8}{@{}l}{cost-adjusted success, mean6 $-$ 0.05 $\times$ tool calls (/ also charging judgment calls)} \\
\qtwo & 0.010 & 0.053 & 0.083 & 0.014 & 0.078 & 0.090 / $-$0.053 & 0.173 / $-$0.027 \\
\llama & $-$0.043 & 0.019 & 0.043 & 0.016 & 0.054 & 0.062 / $-$0.192 & 0.093 / $-$0.149 \\
\qthree & 0.208 & 0.285 & 0.272 & 0.252 & 0.263 & 0.269 / 0.069 & 0.304 / 0.130 \\
\qthirty & 0.271 & 0.366 & 0.399 & 0.362 & 0.398 & 0.367 / 0.165 & 0.401 / 0.224 \\
\bottomrule
\end{tabular}\end{adjustbox}
\caption{Stopping measures and cost for the arms with own judgments, HotpotQA test300, raw analysis. Design-label contrast for other arms and models: thinking $-0.02$ [$-0.13, +0.08$] (\qthree), \sonnet\ unaided $+0.12$ [$+0.05, +0.18$], \haiku\ unaided $-0.02$ [$-0.04, -0.01$].}
\label{tab:deltaown}
\end{table*}

\begin{table*}[tp]
\centering
\begin{minipage}[t]{0.48\textwidth}\vspace{0pt}\centering
\footnotesize\setlength{\tabcolsep}{3pt}
\begin{adjustbox}{max width=\linewidth}\begin{tabular}{@{}lccc@{}}
\toprule
Simulated policy & \ii & design $\Delta$ & own $\Delta$ \\
\midrule
never stops & +0.00 & +0.00 & +0.00 \\
answers at the deadline & +0.00 & +0.00 & +0.00 \\
clock, action 4 & -- & +0.00 & +0.00 \\
clock, action 6 & +0.50 & +0.00 & +0.00 \\
hazard rising with time & +0.18 & $-$0.02 & 0.00 \\
own-judgment count $k=3$ & +0.39 & +0.60 & +1.00 \\
own-judgment count $k=5$ & +0.44 & +0.21 & +0.38 \\
own-judgment count $k=7$ & +0.00 & +0.00 & +0.00 \\
design-label count $k=3$ & -- & +1.00 & +0.33 \\
design-label count $k=5$ & +0.37 & +0.33 & +0.20 \\
hazard rising with own count & +0.32 & +0.18 & +0.31 \\
\bottomrule
\end{tabular}\end{adjustbox}
\caption{The stopping measures on simulated policies with known behavior.}
\label{tab:sim}
\end{minipage}\hfill
\begin{minipage}[t]{0.48\textwidth}\vspace{0pt}\centering
\footnotesize\setlength{\tabcolsep}{2.5pt}
\begin{adjustbox}{max width=\linewidth}\begin{tabular}{@{}lcccccc@{}}
\toprule
& judged & answers after & \multicolumn{2}{c}{$\Delta$ (own)} & \multicolumn{2}{c}{success} \\
\cmidrule(lr){4-5}\cmidrule(l){6-7}
Model & useless & 5 useless & none & rule & none & rule \\
\midrule
\qtwo & 97.5\% & 39/200 & $-0.02$ & $+0.16$ & 0.077 & 0.160 \\
\llama & 87.9\% & 5/207 & $-0.01$ & $+0.45$ & 0.047 & 0.140 \\
\qthree & 86.5\% & 26/176 & $-0.03$ & $+0.35$ & 0.120 & 0.190 \\
\qthirty & 84.3\% & 40/174 & $-0.10$ & $+0.25$ & 0.113 & 0.233 \\
\bottomrule
\end{tabular}\end{adjustbox}
\caption{Plausible regime, HotpotQA test300, raw analysis.}
\label{tab:plausible}
\end{minipage}
\end{table*}

\paragraph{Plausible regime.} Here every failed observation is one of the question's own distractor paragraphs, which are on topic. For the unaided agent, with judgments replayed on its own trajectories, Table~\ref{tab:plausible} gives the share of these paragraphs its judgments call useless, how often it answers once its judgments reach five consecutive ``useless'', and the own-judgment contrast within this regime. It also gives the rule arm's contrast and both arms' success (exploratory).

\paragraph{Decide (exploratory).} $\Delta$ on the judgments recorded during the decide runs (Figure~\ref{fig:delta}): \qtwo $-0.24$ [$-0.37, -0.13$]; \llama $-0.14$ [$-0.18, -0.11$]; \qthree $-0.15$ [$-0.29, -0.01$]. Decide was not run for \qthirty.

\subsection{Stopping hazards and the integration index}\label{app:hazard}
\begin{figure*}[tp]
\centering
\includegraphics[width=\textwidth]{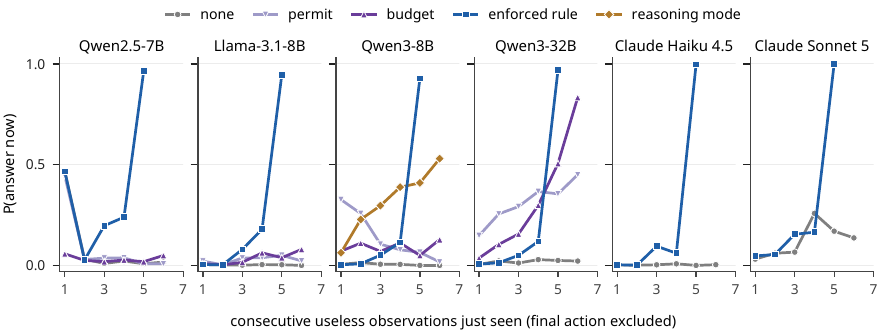}
\caption{Stopping hazard: probability of answering at a decision point after $j$ consecutive failed observations (final-action decisions excluded; 8-action budget; test300, \sonnet on 100 questions). Unaided agents (gray) do not rise with $j$, except for a partial rise in \sonnet. Prompt cues (purple) stay flat or fall in the 7--8B models and rise only in \qthirty. The enforced rule (blue) steps up at $j=5$; the reasoning mode (brown, \qthree) and the budget-cued \qthirty rise with $j$, but not with the evidence at a fixed step (\qthirty: $\Delta=0.00$, Figure~\ref{fig:delta}; thinking: design-label contrast $-0.02$, Appendix~\ref{app:add12}).}
\label{fig:hazard}
\end{figure*}

Figure~\ref{fig:hazard} gives the stopping hazard by the number of consecutive failed observations just seen, and Table~\ref{tab:index} the pre-registered integration index for the arms in Table~\ref{tab:main}.

\begin{table*}[tp]
\centering\footnotesize\setlength{\tabcolsep}{4pt}
\begin{tabular}{@{}l cccccccc@{}}
\toprule
Model & none & permit & budget & stated & cost & decide & enforced & combo \\
\midrule
\qtwo & ${-}.32/{-}.11$ & ${-}.29/{-}.16$ & ${-}.02/{-}.18$ & ${-}.18/{-}.31$ & $.00/{-}.18$ & ${-}.33/{-}.24$ & $.18/.33$ & $.47/.21$ \\
\llama & $.00/{-}.02$ & $.03/{-}.19$ & $.05/{-}.10$ & $.31/.09$ & $.02/{-}.17$ & $.05/{-}.14$ & $.48/.37$ & $.49/.27$ \\
\qthree & ${-}.01/{-}.06$ & ${-}.23/{-}.32$ & $.00/{-}.20$ & $.03/{-}.18$ & $.00/{-}.19$ & ${-}.29/{-}.15$ & $.47/.33$ & $.43/.20$ \\
\qthirty & $.01/{-}.07$ & $.17/.01$ & $.32/.00$ & $.54/.08$ & $.41/.09$ & -- & $.49/.33$ & $.52/.19$ \\
\haiku & $.00/\text{--}$ & -- & -- & -- & -- & -- & $.50/.40$ & -- \\
\sonnet$^\dagger$ & $.18/\text{--}$ & -- & -- & -- & -- & -- & $.49/\text{--}$ & -- \\
\bottomrule
\end{tabular}
\caption{Integration index \ii\ (pre-registered for some arms, Appendix~\ref{app:prereg}) and time-matched contrast $\Delta$ on the agent's own judgments, test300, raw analysis. $^\dagger$First 100 questions; its per-step judgments were not saved (Appendix~\ref{app:sonnet}).}
\label{tab:index}
\end{table*}

\paragraph{Hazard model.} $\Delta$ compares histories with and without an earlier result judged useful, so a negative $\Delta$ could come from answering on sufficient evidence rather than from ignoring useless evidence. Table~\ref{tab:hazmodel} separates the two with a discrete-time linear probability model on every decision after a result the agent judged useless (decisions after one to six results, the five failure regimes): answering $= \alpha_{\text{action}} + \alpha_{\text{regime}} + \beta\cdot\text{run} + \gamma\cdot\text{earlier useful} + \varepsilon$, where run is the current number of consecutive useless judgments. $\beta>0$ means stopping rises with accumulated useless evidence; $\gamma>0$ means stopping follows earlier useful evidence. Without an earlier useful judgment, run equals the number of results so far, so $\beta$ is identified from histories with one. The second line adds question fixed effects (within-question estimator). Each fit uses 3,064--6,434 decisions, 5--31\% of them with an earlier useful judgment. Intervals: question-cluster bootstrap, 1,000 resamples. Exploratory.

\begin{table*}[tp]
\centering\footnotesize\setlength{\tabcolsep}{5pt}
\begin{adjustbox}{max width=\linewidth}\begin{tabular}{@{}l ccccccc@{}}
\toprule
Model & none & permit & budget & stated & cost & enforced & combo \\
\midrule
\multicolumn{8}{@{}l}{\emph{test300}: $\beta$ (run of useless judgments), pooled; second line with question fixed effects} \\
\qtwo & \makecell{$-.01$\\[-1pt]{\scriptsize[$-.04, +.03$]}} & \makecell{$+.01$\\[-1pt]{\scriptsize[$-.06, +.07$]}} & \makecell{$-.03$\\[-1pt]{\scriptsize[$-.08, +.01$]}} & \makecell{$-.09$\\[-1pt]{\scriptsize[$-.27, +.01$]}} & \makecell{$-.05$\\[-1pt]{\scriptsize[$-.09, -.02$]}} & \makecell{$+.17$\\[-1pt]{\scriptsize[$+.14, +.21$]}} & \makecell{$+.13$\\[-1pt]{\scriptsize[$+.09, +.17$]}} \\
 & \makecell{$-.00$\\[-1pt]{\scriptsize[$-.05, +.05$]}} & \makecell{$+.00$\\[-1pt]{\scriptsize[$-.06, +.07$]}} & \makecell{$-.05$\\[-1pt]{\scriptsize[$-.11, -.00$]}} & \makecell{$-.09$\\[-1pt]{\scriptsize[$-.31, +.03$]}} & \makecell{$-.05$\\[-1pt]{\scriptsize[$-.09, -.01$]}} & \makecell{$+.18$\\[-1pt]{\scriptsize[$+.12, +.23$]}} & \makecell{$+.11$\\[-1pt]{\scriptsize[$+.07, +.15$]}} \\
\addlinespace[3pt]
\llama & \makecell{$-.01$\\[-1pt]{\scriptsize[$-.01, +.00$]}} & \makecell{$-.04$\\[-1pt]{\scriptsize[$-.07, -.02$]}} & \makecell{$-.03$\\[-1pt]{\scriptsize[$-.05, -.01$]}} & \makecell{$+.04$\\[-1pt]{\scriptsize[$+.02, +.06$]}} & \makecell{$-.05$\\[-1pt]{\scriptsize[$-.07, -.03$]}} & \makecell{$+.15$\\[-1pt]{\scriptsize[$+.14, +.16$]}} & \makecell{$+.11$\\[-1pt]{\scriptsize[$+.09, +.13$]}} \\
 & \makecell{$-.01$\\[-1pt]{\scriptsize[$-.02, .00$]}} & \makecell{$-.04$\\[-1pt]{\scriptsize[$-.06, -.01$]}} & \makecell{$-.02$\\[-1pt]{\scriptsize[$-.04, -.01$]}} & \makecell{$+.04$\\[-1pt]{\scriptsize[$+.02, +.07$]}} & \makecell{$-.05$\\[-1pt]{\scriptsize[$-.07, -.03$]}} & \makecell{$+.14$\\[-1pt]{\scriptsize[$+.13, +.16$]}} & \makecell{$+.10$\\[-1pt]{\scriptsize[$+.08, +.13$]}} \\
\addlinespace[3pt]
\qthree & \makecell{$-.02$\\[-1pt]{\scriptsize[$-.03, +.00$]}} & \makecell{$-.07$\\[-1pt]{\scriptsize[$-.12, +.00$]}} & \makecell{$-.04$\\[-1pt]{\scriptsize[$-.07, +.00$]}} & \makecell{$-.03$\\[-1pt]{\scriptsize[$-.08, +.02$]}} & \makecell{$-.05$\\[-1pt]{\scriptsize[$-.08, -.01$]}} & \makecell{$+.15$\\[-1pt]{\scriptsize[$+.13, +.17$]}} & \makecell{$+.11$\\[-1pt]{\scriptsize[$+.08, +.15$]}} \\
 & \makecell{$-.01$\\[-1pt]{\scriptsize[$-.03, +.01$]}} & \makecell{$-.10$\\[-1pt]{\scriptsize[$-.19, -.02$]}} & \makecell{$-.01$\\[-1pt]{\scriptsize[$-.05, +.03$]}} & \makecell{$-.00$\\[-1pt]{\scriptsize[$-.05, +.05$]}} & \makecell{$-.03$\\[-1pt]{\scriptsize[$-.07, +.01$]}} & \makecell{$+.16$\\[-1pt]{\scriptsize[$+.14, +.18$]}} & \makecell{$+.14$\\[-1pt]{\scriptsize[$+.09, +.18$]}} \\
\addlinespace[3pt]
\qthirty & \makecell{$-.04$\\[-1pt]{\scriptsize[$-.06, -.02$]}} & \makecell{$+.00$\\[-1pt]{\scriptsize[$-.06, +.07$]}} & \makecell{$+.04$\\[-1pt]{\scriptsize[$+.01, +.07$]}} & \makecell{$+.07$\\[-1pt]{\scriptsize[$+.03, +.12$]}} & \makecell{$+.03$\\[-1pt]{\scriptsize[$-.01, +.06$]}} & \makecell{$+.14$\\[-1pt]{\scriptsize[$+.13, +.16$]}} & \makecell{$+.09$\\[-1pt]{\scriptsize[$+.05, +.14$]}} \\
 & \makecell{$-.03$\\[-1pt]{\scriptsize[$-.05, -.02$]}} & \makecell{$-.01$\\[-1pt]{\scriptsize[$-.09, +.06$]}} & \makecell{$+.05$\\[-1pt]{\scriptsize[$+.01, +.09$]}} & \makecell{$+.10$\\[-1pt]{\scriptsize[$+.06, +.15$]}} & \makecell{$+.05$\\[-1pt]{\scriptsize[$+.01, +.09$]}} & \makecell{$+.15$\\[-1pt]{\scriptsize[$+.13, +.18$]}} & \makecell{$+.11$\\[-1pt]{\scriptsize[$+.07, +.16$]}} \\
\midrule
\multicolumn{8}{@{}l}{\emph{fresh300}: $\beta$ (run of useless judgments), pooled; second line with question fixed effects} \\
\qtwo & \makecell{$-.01$\\[-1pt]{\scriptsize[$-.05, +.02$]}} & -- & \makecell{$-.03$\\[-1pt]{\scriptsize[$-.07, +.00$]}} & -- & -- & \makecell{$+.17$\\[-1pt]{\scriptsize[$+.13, +.21$]}} & -- \\
 & \makecell{$-.06$\\[-1pt]{\scriptsize[$-.12, -.01$]}} &  & \makecell{$-.04$\\[-1pt]{\scriptsize[$-.08, .00$]}} &  &  & \makecell{$+.13$\\[-1pt]{\scriptsize[$+.07, +.18$]}} &  \\
\addlinespace[3pt]
\llama & \makecell{$-.01$\\[-1pt]{\scriptsize[$-.02, -.00$]}} & -- & \makecell{$-.05$\\[-1pt]{\scriptsize[$-.06, -.03$]}} & -- & -- & \makecell{$+.15$\\[-1pt]{\scriptsize[$+.14, +.16$]}} & -- \\
 & \makecell{$-.01$\\[-1pt]{\scriptsize[$-.02, -.00$]}} &  & \makecell{$-.05$\\[-1pt]{\scriptsize[$-.07, -.03$]}} &  &  & \makecell{$+.14$\\[-1pt]{\scriptsize[$+.13, +.16$]}} &  \\
\addlinespace[3pt]
\qthree & \makecell{$.00$\\[-1pt]{\scriptsize[$-.01, +.01$]}} & -- & \makecell{$+.01$\\[-1pt]{\scriptsize[$-.03, +.04$]}} & -- & -- & \makecell{$+.15$\\[-1pt]{\scriptsize[$+.13, +.17$]}} & -- \\
 & \makecell{$+.00$\\[-1pt]{\scriptsize[$-.01, +.01$]}} &  & \makecell{$+.01$\\[-1pt]{\scriptsize[$-.03, +.05$]}} &  &  & \makecell{$+.15$\\[-1pt]{\scriptsize[$+.13, +.17$]}} &  \\
\addlinespace[3pt]
\qthirty & \makecell{$-.02$\\[-1pt]{\scriptsize[$-.03, .00$]}} & -- & \makecell{$+.00$\\[-1pt]{\scriptsize[$-.03, +.04$]}} & -- & -- & \makecell{$+.16$\\[-1pt]{\scriptsize[$+.14, +.17$]}} & -- \\
 & \makecell{$-.01$\\[-1pt]{\scriptsize[$-.03, +.00$]}} &  & \makecell{$+.01$\\[-1pt]{\scriptsize[$-.03, +.04$]}} &  &  & \makecell{$+.16$\\[-1pt]{\scriptsize[$+.14, +.18$]}} &  \\
\midrule
\multicolumn{8}{@{}l}{$\gamma$ (an earlier result judged useful), test300, pooled} \\
\qtwo & \makecell{$+.08$\\[-1pt]{\scriptsize[$+.02, +.16$]}} & \makecell{$+.19$\\[-1pt]{\scriptsize[$+.08, +.33$]}} & \makecell{$+.06$\\[-1pt]{\scriptsize[$-.02, +.16$]}} & \makecell{$-.00$\\[-1pt]{\scriptsize[$-.20, +.13$]}} & \makecell{$+.02$\\[-1pt]{\scriptsize[$-.05, +.09$]}} & \makecell{$+.11$\\[-1pt]{\scriptsize[$+.04, +.18$]}} & \makecell{$+.08$\\[-1pt]{\scriptsize[$-.01, +.17$]}} \\
\llama & \makecell{$+.01$\\[-1pt]{\scriptsize[$-.01, +.02$]}} & \makecell{$+.04$\\[-1pt]{\scriptsize[$-.03, +.10$]}} & \makecell{$+.00$\\[-1pt]{\scriptsize[$-.04, +.05$]}} & \makecell{$+.02$\\[-1pt]{\scriptsize[$-.04, +.07$]}} & \makecell{$+.01$\\[-1pt]{\scriptsize[$-.04, +.06$]}} & \makecell{$+.01$\\[-1pt]{\scriptsize[$-.02, +.04$]}} & \makecell{$+.01$\\[-1pt]{\scriptsize[$-.03, +.06$]}} \\
\qthree & \makecell{$+.01$\\[-1pt]{\scriptsize[$-.02, +.06$]}} & \makecell{$+.10$\\[-1pt]{\scriptsize[$-.05, +.31$]}} & \makecell{$+.05$\\[-1pt]{\scriptsize[$-.03, +.15$]}} & \makecell{$+.03$\\[-1pt]{\scriptsize[$-.08, +.18$]}} & \makecell{$+.02$\\[-1pt]{\scriptsize[$-.06, +.12$]}} & \makecell{$+.06$\\[-1pt]{\scriptsize[$+.01, +.11$]}} & \makecell{$+.07$\\[-1pt]{\scriptsize[$-.03, +.19$]}} \\
\qthirty & \makecell{$-.04$\\[-1pt]{\scriptsize[$-.08, -.01$]}} & \makecell{$-.03$\\[-1pt]{\scriptsize[$-.20, +.16$]}} & \makecell{$+.11$\\[-1pt]{\scriptsize[$+.03, +.20$]}} & \makecell{$+.11$\\[-1pt]{\scriptsize[$-.01, +.23$]}} & \makecell{$-.04$\\[-1pt]{\scriptsize[$-.12, +.05$]}} & \makecell{$+.02$\\[-1pt]{\scriptsize[$-.02, +.07$]}} & \makecell{$+.01$\\[-1pt]{\scriptsize[$-.07, +.12$]}} \\
\bottomrule
\end{tabular}\end{adjustbox}
\caption{Hazard model: change in the probability of answering per additional consecutive useless judgment ($\beta$) and after an earlier useful judgment ($\gamma$), with 95\% intervals.}
\label{tab:hazmodel}
\end{table*}

\Needspace*{16\baselineskip}
\subsection{Specificity of side-channel judgments}\label{app:spec}
\noindent\begin{minipage}[t]{0.44\textwidth}\vspace{0pt}
Table~\ref{tab:spec} gives the share of side-channel judgments in the rule arm's HotpotQA test300 log that call a \emph{real} search result useful. It splits them by whether the page is one of the question's two gold paragraphs or one of its eight distractor paragraphs, and by whether the agent has already seen it in the trajectory. Real distractor paragraphs are usually useless for the question, and a repeated page adds nothing new, so low rates there are appropriate.

\end{minipage}\hfill
\begin{minipage}[t]{0.53\textwidth}\vspace{0pt}
\centering\footnotesize\setlength{\tabcolsep}{3pt}
\begin{adjustbox}{max width=\linewidth}\begin{tabular}{@{}lcccc@{}}
\toprule
& \multicolumn{2}{c}{gold paragraph} & \multicolumn{2}{c}{distractor paragraph} \\
\cmidrule(lr){2-3}\cmidrule(l){4-5}
Model & first & repeat & first & repeat \\
\midrule
\qtwo & 0.57 (1045) & 0.03 (76) & 0.09 (442) & 0.00 (235) \\
\llama & 0.92 (1771) & 0.09 (150) & 0.41 (689) & 0.05 (438) \\
\qthree & 0.96 (1817) & 0.17 (155) & 0.36 (745) & 0.08 (825) \\
\qthirty & 0.98 (2026) & 0.06 (194) & 0.38 (849) & 0.03 (703) \\
\haiku & 0.77 (2157) & 0.05 (274) & 0.08 (884) & 0.00 (365) \\
\bottomrule
\end{tabular}\end{adjustbox}
\captionof{table}{Share of side-channel USEFUL judgments on real search results (number of judgments in parentheses).}
\label{tab:spec}
\end{minipage}

\subsection{Beliefs probe}\label{app:beliefs}

\paragraph{Probe.} At every decision point of a recorded trajectory we fork the history, append one assessment turn, and discard the reply; nothing is written back. Two turns are asked separately, each on its own fork of the same history: a usefulness turn (five categories aligned with the annotation rubric) and a continuation turn that asks for $p_\text{continue}$ and $p_\text{answer\_now}$. The continuation turn does not state the remaining budget (``an unspecified number of tool interactions''), offers no action, and never mentions the action syntax. Replies that do not parse are recorded as missing, never coerced.

\begin{figure*}[tp]
\centering
\includegraphics[width=\textwidth]{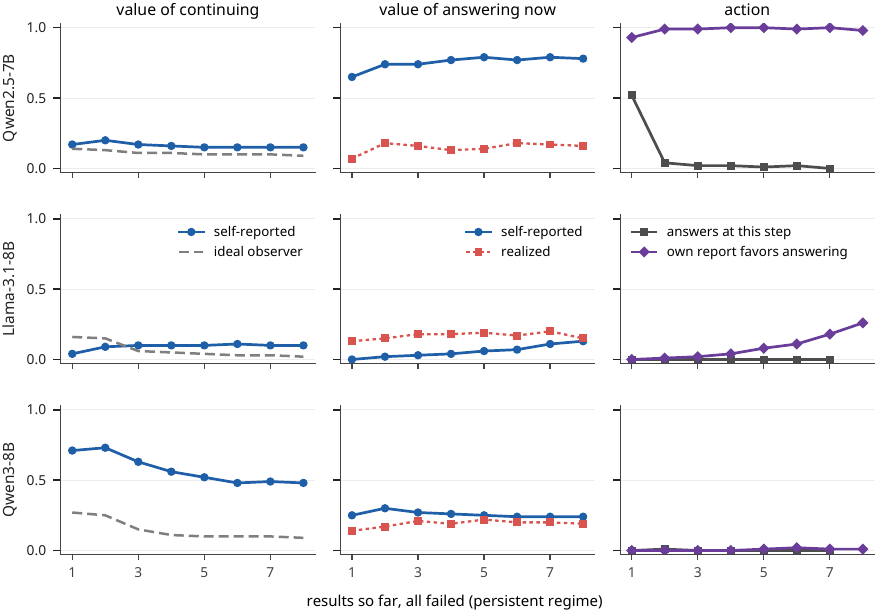}
\caption{The three models err in different directions, and where a model's own report favors answering, its action does not follow. Held-out persistent regime, placeholder answers repaired; means over decision points, by the number of failed results so far. \textbf{Left:} self-reported value of continuing against the ideal observer (the realized value is the success on this source, 0.00--0.04). \textbf{Middle:} self-reported value of answering now against the realized value of the answer-now branch. \textbf{Right:} share of decision points at which the agent answers, and at which its own report favors answering.}
\label{fig:beliefs}
\end{figure*}

\paragraph{Realized values.} $V_c$ is the success of the recorded trajectory from that decision point on; $V_f$ is the success of an answer-now branch: the same history replayed with the rule's finish instruction (Appendix~\ref{app:prompts}), which the agents obey at 89--98\% of decision points. Both are computed on the unaided test runs of \qtwo, \llama and \qthree in four regimes; this analysis is descriptive, not pre-registered. The ideal observer $E[V_c \mid k]$ after $k$ results that all failed is a reference rather than an optimum. It puts an equal prior on the four regimes the probe was run on (clean, persistent, plausible, recover\_after\_2) and updates it with the rate at which the model's own usefulness replies call the result at each position in each regime not newly useful. It then averages the regimes' realized values of continuing. Figure~\ref{fig:beliefs} shows the three quantities for the three models on which the probe was run.

\subsection{Answer repair}\label{app:repair}

\paragraph{What happened.} After the runs we found answers that copied the template \texttt{finish[answer]}: the word ``answer'' (sometimes ``guess'') or no argument, scored wrong. The forced-answer wording and the prompt arms' ``use finish[answer] with your best guess'' elicited them most. Placeholders can only lower success, and they occurred almost only in arms that stop and answer, so H1 and H2 were conservative. Contrasts between answering arms could move either way, and stopping measures are unaffected.

\paragraph{Procedure (post hoc, identical for all arms).} For every answered record whose answer was empty or a template string (``answer'', ``guess'', ``your best guess'' and six similar), in every arm and model except the call-cost arm, the fresh-question runs, the third run and the runs of amendments 13--20 (which are reported raw only), we rebuilt the exact conversation, kept the model's finish turn, and appended one fixed message: ``Your Action Input was a placeholder, not an answer. Respond again in the required format (Thought / Action / Action Input) with Action: finish and your actual final answer to the original question as the Action Input.'' One attempt, unchanged decoding; a second placeholder stayed wrong. Only answers and scores changed. Raw results are primary; repaired ones are a sensitivity analysis, used only for the realized values of the belief probe (Appendix~\ref{app:beliefs}).

\paragraph{Frequency.} On HotpotQA test300 (2,100 trajectories per arm), placeholders concentrated in \qthree's stated-rule, budget, combination, rule, lexical and permit arms (297, 231, 226, 188, 146, 114), \llama's stated-rule arm (149) and \qtwo's rule and random arms (105, 82); the unaided arms had one in total, and the unrepaired call-cost arm 364 for \qthree. Including 2,628 in the answer-now branch, the local models produced 3,834 (stated-rule and combination arms excluded); 3,254 became real answers, 545 correct; \qtwo complied least (rule arm: 42 of 105). \haiku had 13, \qthirty 24, \sonnet none; FEVER's seven original arms had 606, 339 of them empty answers from \llama's decide arm.

\begin{table*}[tp]
\centering\footnotesize
\setlength{\tabcolsep}{2.1pt}
\begin{tabular}{@{}l rrr rrr rrr rrr@{}}
\toprule
& \multicolumn{6}{c}{HotpotQA test300} & \multicolumn{6}{c}{FEVER} \\
\cmidrule(lr){2-7}\cmidrule(l){8-13}
& \multicolumn{3}{c}{raw} & \multicolumn{3}{c}{repaired} & \multicolumn{3}{c}{raw} & \multicolumn{3}{c}{repaired} \\
\cmidrule(lr){2-4}\cmidrule(lr){5-7}\cmidrule(lr){8-10}\cmidrule(l){11-13}
Contrast (mean6) & Q2.5 & L & Q3 & Q2.5 & L & Q3 & Q2.5 & L & Q3 & Q2.5 & L & Q3 \\
\midrule
rule $-$ none (H1, F-H1)      & $+.045$ & $+.064$ & $+.026$ & $+.046$ & $+.069$ & $+.040$ & $+.129$ & $+.201$ & $+.142$ & $+.134$ & $+.203$ & $+.142$ \\
rule $-$ random (H2, F-H2)    & $+.011$ & $+.029$ & $+.021$ & $+.011$ & $+.032$ & $+.033$ & $+.044$ & $+.123$ & $+.096$ & $+.045$ & $+.124$ & $+.096$ \\
rule $-$ decide (H4)          & $-.123$ & $+.062$ & $+.048$ & $-.123$ & $+.059$ & $+.062$ & $-.084$ & $+.350$ & $+.005$ & $-.079$ & $+.180$ & $+.005$ \\
rule $-$ permit (H7)          & $+.023$ & $+.059$ & $+.051$ & $+.024$ & $+.061$ & $+.054$ & $+.138$ & $+.138$ & $+.064$ & $+.142$ & $+.108$ & $+.062$ \\
rule $-$ budget (H8)          & $-.091$ & $+.016$ & $+.009$ & $-.091$ & $+.019$ & $-.011$ & $-.024$ & $+.066^\dagger$ & $-.021^\dagger$ & $-.022$ & $+.022^\dagger$ & $-.026^\dagger$ \\
combo $-$ rule (C1)           & $+.138$ & $+.009$ & $+.004$ & $+.139$ & $+.007$ & $+.024$ & $+.088$ & $+.026$ & $+.051$ & $+.093$ & $+.064$ & $+.053$ \\
combo $-$ budget (C1)         & $+.047$ & $+.025$ & $+.013^\dagger$ & $+.048$ & $+.026$ & $+.013^\dagger$ & $+.064$ & $+.091$ & $+.030$ & $+.071$ & $+.086$ & $+.027$ \\
\bottomrule
\end{tabular}
\caption{Key contrasts, raw vs.\ repaired (IDs refer to HotpotQA; FEVER H4/H7/H8 counterparts are exploratory). Q2.5: \qtwo; L: \llama; Q3: \qthree. $^\dagger$Significance (95\% interval excluding zero) differs between analyses. Also unchanged in decision: H1 for \haiku ($+.097$ both), L-H1 ($+.063$ both), \qthree's H3 lower bounds on rec1/rec3 ($-.057$/$-.060$ raw, $-.057$/$-.057$ repaired), H6 ($+.003$ raw, $-.011$ repaired, both n.s.). Realized answer-now success on persistent: $.12$/$.15$/$.14$ raw, $.14$/$.17$/$.19$ repaired.}
\label{tab:repair}
\end{table*}

\paragraph{Effect.} No decision in Table~\ref{tab:repair} changed (in the second run, Rp4 for \qthree\ passes only after repair; Appendix~\ref{app:rep}); all failures remain (\qtwo's H7 at $p=0.0505$). Three contrasts in the table changed significance, as did \qthree's stated-rule contrast in PP2 (Table~\ref{tab:prereg}). On FEVER, the rule's lead over the budget cue for \llama fell from $+0.066$ [$+0.041$, $+0.090$] to $+0.022$ [$-0.005$, $+0.047$], and the budget cue's lead for \qthree became significant ($-0.026$ [$-0.048$, $-0.003$]). On HotpotQA, the combination's lead over the budget cue for \qthree moved from [$-0.000$, $+0.028$] to [$+0.001$, $+0.026$]. \llama's FEVER decide arm rose from 0.401 to 0.574; the unaided arm scores 0.551.

\section{Additional results}\label{app:results}

\subsection{Full result tables}\label{app:tables}

All numbers in this appendix are computed by one script from the run logs. HotpotQA results use test300 with the 8-action budget (\sonnet: its first 100 questions); FEVER results use all 292 claims, for which no \emph{plausible} regime was run. Success is the share of questions answered correctly in a regime; mean6 averages the six regimes of \S\ref{sec:setting} with equal weight. Unless a column says \emph{repaired}, numbers come from the raw, pre-registered analysis, in which a template-placeholder answer counts as wrong (Appendix~\ref{app:repair}). Intervals are 95\% percentile intervals over 2,000 bootstrap resamples of questions (for \ii, whole questions are resampled with all their decision points); the one-sided $p$ is the share of resampled differences $\le 0$, unadjusted for multiplicity. Condition names are as in Table~\ref{tab:main}, with short forms: rule = enforced rule, policy or stated = stated rule, cost = call cost, think = reasoning mode, random and lexical = the enforced rule on a random or lexical signal; ``--'' marks a condition or regime that was not run. Table~\ref{tab:app-hotpot} gives per-regime success and the integration index on HotpotQA, Table~\ref{tab:app-fever} the same on FEVER, Table~\ref{tab:app-contrasts} the paired contrasts, and Table~\ref{tab:app-budget16} the budget-scaling runs.

\begin{table*}[tp]
\centering\footnotesize
\setlength{\tabcolsep}{3.2pt}
\begin{tabular}{@{}ll ccccccc c cccc@{}}
\toprule
 & & \multicolumn{7}{c}{success per regime} & & \multicolumn{4}{c}{integration index} \\
\cmidrule(lr){3-9}\cmidrule(l){11-14}
Model & Arm & pers. & rec1 & rec2 & rec3 & late & clean & plaus. & mean6 & \ii & 95\% CI & $h_{4\text{-}5}$ & $h_{1\text{-}2}$ \\
\midrule
\qtwo & none & .043 & .160 & .153 & .147 & .220 & .427 & .077 & .192 & $-.32$ & $[-.37, -.27]$ & .02 & .34 \\
 & permit & .033 & .170 & .207 & .137 & .303 & .430 & .143 & .213 & $-.29$ & $[-.34, -.24]$ & .03 & .32 \\
 & budget & .090 & .407 & .377 & .327 & .333 & .433 & .183 & .328 & $-.02$ & $[-.04, .00]$ & .02 & .04 \\
 & decide & .183 & .370 & .300 & .277 & .517 & .513 & .230 & .360 & $-.33$ & $[-.37, -.29]$ & .03 & .36 \\
 & random & .080 & .187 & .183 & .163 & .300 & .443 & .143 & .226 & $-.11$ & $[-.17, -.05]$ & .23 & .33 \\
 & lexical & .090 & .173 & .170 & .160 & .313 & .440 & .110 & .224 & .15 & $[.09, .20]$ & .48 & .33 \\
 & rule & .100 & .193 & .173 & .163 & .327 & .463 & .160 & .237 & .18 & $[.13, .22]$ & .51 & .33 \\
 & combo & .163 & .417 & .420 & .370 & .393 & .487 & .243 & .375 & .47 & $[.45, .49]$ & .51 & .04 \\
\midrule
\llama & none & .000 & .370 & .300 & .280 & .207 & .423 & .047 & .263 & .00 & $[-.01, .01]$ & .00 & .00 \\
 & permit & .017 & .363 & .310 & .270 & .263 & .390 & .090 & .269 & .03 & $[.01, .05]$ & .04 & .01 \\
 & budget & .057 & .390 & .353 & .307 & .323 & .440 & .137 & .312 & .05 & $[.03, .06]$ & .05 & .01 \\
 & decide & .033 & .343 & .303 & .223 & .287 & .403 & .113 & .266 & .05 & $[.03, .07]$ & .07 & .02 \\
 & random & .033 & .387 & .337 & .293 & .267 & .473 & .093 & .298 & .09 & $[.07, .11]$ & .09 & .00 \\
 & lexical & .160 & .370 & .333 & .260 & .377 & .470 & .060 & .328 & .44 & $[.43, .46]$ & .45 & .00 \\
 & rule & .143 & .373 & .327 & .273 & .377 & .473 & .140 & .328 & .48 & $[.46, .50]$ & .49 & .00 \\
 & combo & .177 & .390 & .330 & .303 & .400 & .420 & .160 & .337 & .49 & $[.47, .51]$ & .50 & .01 \\
\midrule
\qthree & none & .007 & .553 & .537 & .500 & .487 & .603 & .120 & .448 & $-.01$ & $[-.01, .00]$ & .00 & .01 \\
 & permit & .200 & .427 & .393 & .367 & .567 & .587 & .263 & .423 & $-.23$ & $[-.28, -.18]$ & .07 & .30 \\
 & budget & .203 & .503 & .497 & .470 & .537 & .577 & .230 & .464 & .00 & $[-.04, .03]$ & .08 & .09 \\
 & decide & .263 & .503 & .340 & .283 & .570 & .593 & .260 & .426 & $-.29$ & $[-.37, -.19]$ & .13 & .42 \\
 & think & .297 & .533 & .493 & .437 & .543 & .560 & .273 & .477 & .26 & $[.20, .33]$ & .40 & .13 \\
 & random & .050 & .543 & .527 & .487 & .497 & .617 & .130 & .453 & .06 & $[.04, .08]$ & .07 & .01 \\
 & lexical & .167 & .557 & .510 & .510 & .553 & .610 & .140 & .484 & .44 & $[.42, .46]$ & .44 & .01 \\
 & rule & .153 & .527 & .527 & .470 & .547 & .620 & .190 & .474 & .47 & $[.45, .49]$ & .48 & .01 \\
 & combo & .270 & .530 & .487 & .453 & .543 & .583 & .263 & .478 & .43 & $[.40, .45]$ & .52 & .09 \\
\midrule
\qthirty & none & .027 & .630 & .607 & .570 & .543 & .680 & .113 & .509 & .01 & $[.00, .03]$ & .03 & .01 \\
 & permit & .290 & .533 & .510 & .450 & .627 & .693 & .297 & .517 & .17 & $[.11, .23]$ & .36 & .19 \\
 & budget & .330 & .643 & .623 & .540 & .650 & .700 & .337 & .581 & .32 & $[.28, .36]$ & .38 & .07 \\
 & rule & .267 & .647 & .620 & .567 & .643 & .690 & .233 & .572 & .49 & $[.47, .51]$ & .50 & .01 \\
 & combo & .310 & .650 & .597 & .567 & .650 & .693 & .323 & .578 & .52 & $[.50, .55]$ & .59 & .07 \\
\midrule
\haiku & none & .000 & .633 & .623 & .590 & .523 & .637 & .057 & .501 & .00 & $[.00, .01]$ & .00 & .00 \\
 & rule & .380 & .667 & .643 & .577 & .653 & .667 & .363 & .598 & .50 & $[.49, .51]$ & .51 & .00 \\
\midrule
\sonnet & none & .400 & .700 & .670 & .660 & .650 & .710 & -- & .632 & .18 & $[.11, .26]$ & .23 & .04 \\
 & rule & .510 & .700 & .690 & .630 & .690 & .700 & -- & .653 & .49 & $[.46, .52]$ & .54 & .05 \\
\bottomrule
\end{tabular}
\caption{HotpotQA (test300, 8-action budget), raw analysis: success per regime, mean6, and the integration index with its two components, for every model and arm. $n=300$ questions per arm (\sonnet: the first 100; its plausible regime was not run). pers.: persistent; rec$c$: recover\_after\_$c$; late: late\_onset\_from\_3; plaus.: plausible (not part of mean6). \ii $=h_{4\text{-}5}-h_{1\text{-}2}$, where $h_{a\text{-}b}$ is the probability of answering at a decision point just after $a$ to $b$ consecutive failed observations, pooled over the five regimes with failures, decisions at the final action excluded; its 95\% CI is from a question-cluster bootstrap.}
\label{tab:app-hotpot}
\end{table*}

\begin{table*}[tp]
\centering\footnotesize
\setlength{\tabcolsep}{3.2pt}
\begin{tabular}{@{}ll cccccc c cccc@{}}
\toprule
 & & \multicolumn{6}{c}{success per regime} & & \multicolumn{4}{c}{integration index} \\
\cmidrule(lr){3-8}\cmidrule(l){10-13}
Model & Arm & pers. & rec1 & rec2 & rec3 & late & clean & mean6 & \ii & 95\% CI & $h_{4\text{-}5}$ & $h_{1\text{-}2}$ \\
\midrule
\qtwo & none & .127 & .675 & .664 & .644 & .572 & .716 & .566 & $-.12$ & $[-.15, -.10]$ & .01 & .13 \\
 & permit & .151 & .579 & .541 & .538 & .729 & .805 & .557 & $-.20$ & $[-.25, -.16]$ & .02 & .23 \\
 & budget & .384 & .818 & .764 & .798 & .750 & .801 & .719 & .00 & $[-.02, .02]$ & .03 & .03 \\
 & decide & .668 & .822 & .733 & .729 & .870 & .856 & .780 & $-.26$ & $[-.35, -.15]$ & .17 & .43 \\
 & random & .377 & .723 & .695 & .682 & .699 & .733 & .651 & .12 & $[.08, .16]$ & .25 & .14 \\
 & lexical & .445 & .702 & .705 & .685 & .709 & .740 & .664 & .34 & $[.30, .38]$ & .47 & .14 \\
 & rule & .490 & .729 & .726 & .729 & .743 & .753 & .695 & .37 & $[.34, .40]$ & .50 & .14 \\
 & combo & .568 & .832 & .805 & .812 & .849 & .832 & .783 & .47 & $[.45, .49]$ & .50 & .03 \\
\midrule
\llama & none & .007 & .712 & .695 & .664 & .524 & .702 & .551 & .00 & $[.00, .01]$ & .01 & .00 \\
 & permit & .017 & .777 & .760 & .719 & .640 & .767 & .614 & .00 & $[.00, .01]$ & .01 & .01 \\
 & budget & .195 & .812 & .805 & .805 & .699 & .801 & .686 & .01 & $[.01, .02]$ & .02 & .00 \\
 & decide & .116 & .394 & .548 & .507 & .387 & .455 & .401 & .02 & $[.00, .04]$ & .04 & .02 \\
 & random & .209 & .743 & .743 & .705 & .610 & .764 & .629 & .09 & $[.07, .11]$ & .09 & .00 \\
 & lexical & .613 & .760 & .726 & .729 & .777 & .736 & .724 & .45 & $[.43, .48]$ & .45 & .00 \\
 & rule & .610 & .812 & .719 & .726 & .812 & .832 & .752 & .46 & $[.44, .48]$ & .46 & .00 \\
 & combo & .692 & .805 & .774 & .784 & .784 & .825 & .777 & .51 & $[.49, .53]$ & .51 & .00 \\
\midrule
\qthree & none & .051 & .832 & .812 & .805 & .764 & .842 & .684 & $-.02$ & $[-.03, .00]$ & .00 & .02 \\
 & permit & .212 & .880 & .873 & .849 & .866 & .887 & .761 & .00 & $[-.03, .03]$ & .06 & .06 \\
 & budget & .613 & .887 & .897 & .904 & .887 & .894 & .847 & .04 & $[.02, .06]$ & .05 & .01 \\
 & decide & .736 & .849 & .788 & .771 & .894 & .890 & .821 & $-.22$ & $[-.37, -.03]$ & .31 & .53 \\
 & random & .216 & .853 & .836 & .818 & .812 & .846 & .730 & .07 & $[.04, .09]$ & .09 & .02 \\
 & lexical & .592 & .849 & .839 & .822 & .860 & .842 & .801 & .41 & $[.38, .43]$ & .43 & .02 \\
 & rule & .658 & .873 & .853 & .829 & .873 & .870 & .826 & .49 & $[.47, .50]$ & .51 & .02 \\
 & combo & .774 & .904 & .897 & .880 & .897 & .908 & .877 & .51 & $[.49, .52]$ & .52 & .01 \\
\bottomrule
\end{tabular}
\caption{FEVER (292 claims, 8-action budget, every parameter carried over from HotpotQA), raw analysis: success per regime, mean6 and integration index for every arm. $n=292$ claims per arm; no plausible regime. Columns as in Table~\ref{tab:app-hotpot}.}
\label{tab:app-fever}
\end{table*}

\begin{table*}[tp]
\centering\footnotesize
\setlength{\tabcolsep}{4pt}
\begin{tabular}{@{}ll ccc c ccc@{}}
\toprule
 & & \multicolumn{3}{c}{HotpotQA} & & \multicolumn{3}{c}{FEVER} \\
\cmidrule(lr){3-5}\cmidrule(l){7-9}
Model & Contrast & $\Delta$ mean6 & 95\% CI & $p$ &  & $\Delta$ mean6 & 95\% CI & $p$ \\
\midrule
\qtwo & rule\,$-$\,none & $+.045$ & $[+.029, +.062]$ & $<$.001 &  & $+.129$ & $[+.104, +.154]$ & $<$.001 \\
 & rule\,$-$\,permit & $+.023$ & $[-.005, +.051]$ & .055 &  & $+.138$ & $[+.104, +.172]$ & $<$.001 \\
 & rule\,$-$\,budget & $-.091$ & $[-.124, -.060]$ & 1.000 &  & $-.024$ & $[-.055, +.007]$ & .938 \\
 & rule\,$-$\,random & $+.011$ & $[.000, +.022]$ & .029 &  & $+.044$ & $[+.032, +.057]$ & $<$.001 \\
 & rule\,$-$\,lexical & $+.012$ & $[-.002, +.026]$ & .043 &  & $+.031$ & $[+.019, +.043]$ & $<$.001 \\
 & combo\,$-$\,rule & $+.138$ & $[+.104, +.173]$ & $<$.001 &  & $+.088$ & $[+.059, +.118]$ & $<$.001 \\
 & combo\,$-$\,budget & $+.047$ & $[+.028, +.068]$ & $<$.001 &  & $+.064$ & $[+.046, +.083]$ & $<$.001 \\
\midrule
\llama & rule\,$-$\,none & $+.064$ & $[+.046, +.082]$ & $<$.001 &  & $+.201$ & $[+.178, +.224]$ & $<$.001 \\
 & rule\,$-$\,permit & $+.059$ & $[+.033, +.086]$ & $<$.001 &  & $+.138$ & $[+.115, +.163]$ & $<$.001 \\
 & rule\,$-$\,budget & $+.016$ & $[-.013, +.047]$ & .157 &  & $+.066$ & $[+.041, +.090]$ & $<$.001 \\
 & rule\,$-$\,random & $+.029$ & $[+.013, +.044]$ & $<$.001 &  & $+.123$ & $[+.103, +.142]$ & $<$.001 \\
 & rule\,$-$\,lexical & $-.001$ & $[-.014, +.013]$ & .539 &  & $+.028$ & $[+.012, +.044]$ & $<$.001 \\
 & combo\,$-$\,rule & $+.009$ & $[-.019, +.037]$ & .286 &  & $+.026$ & $[+.001, +.050]$ & .024 \\
 & combo\,$-$\,budget & $+.025$ & $[+.006, +.043]$ & .004 &  & $+.091$ & $[+.070, +.112]$ & $<$.001 \\
\midrule
\qthree & rule\,$-$\,none & $+.026$ & $[+.012, +.041]$ & .001 &  & $+.142$ & $[+.124, +.160]$ & $<$.001 \\
 & rule\,$-$\,permit & $+.051$ & $[+.020, +.079]$ & .001 &  & $+.064$ & $[+.045, +.084]$ & $<$.001 \\
 & rule\,$-$\,budget & $+.009$ & $[-.018, +.037]$ & .249 &  & $-.021$ & $[-.043, +.001]$ & .970 \\
 & rule\,$-$\,random & $+.021$ & $[+.008, +.033]$ & .002 &  & $+.096$ & $[+.082, +.110]$ & $<$.001 \\
 & rule\,$-$\,lexical & $-.011$ & $[-.022, +.001]$ & .963 &  & $+.025$ & $[+.015, +.036]$ & $<$.001 \\
 & combo\,$-$\,rule & $+.004$ & $[-.023, +.032]$ & .389 &  & $+.051$ & $[+.031, +.071]$ & $<$.001 \\
 & combo\,$-$\,budget & $+.013$ & $[.000, +.028]$ & .028 &  & $+.030$ & $[+.017, +.043]$ & $<$.001 \\
\midrule
\qthirty & rule\,$-$\,none & $+.063$ & $[+.045, +.080]$ & $<$.001 &  & -- &  &  \\
 & rule\,$-$\,permit & $+.055$ & $[+.026, +.084]$ & .001 &  & -- &  &  \\
 & rule\,$-$\,budget & $-.009$ & $[-.035, +.017]$ & .744 &  & -- &  &  \\
 & combo\,$-$\,rule & $+.006$ & $[-.022, +.033]$ & .346 &  & -- &  &  \\
 & combo\,$-$\,budget & $-.003$ & $[-.019, +.012]$ & .640 &  & -- &  &  \\
\midrule
\haiku & rule\,$-$\,none & $+.097$ & $[+.080, +.114]$ & $<$.001 &  & -- &  &  \\
\midrule
\sonnet & rule\,$-$\,none & $+.022$ & $[-.003, +.052]$ & .051 &  & -- &  &  \\
\bottomrule
\end{tabular}
\caption{Paired contrasts in mean6 between arms, raw analysis: difference (first arm minus second), 95\% interval over 2,000 question-level bootstrap resamples, and one-sided bootstrap $p$ for a positive difference (unadjusted; the main text applies Holm correction across models where pre-registered). $n=300$ paired questions on HotpotQA (\sonnet: 100) and 292 on FEVER; \qthirty, \haiku and \sonnet were not run on FEVER.}
\label{tab:app-contrasts}
\end{table*}

\begin{table*}[tp]
\centering\footnotesize
\setlength{\tabcolsep}{3.4pt}
\begin{tabular}{@{}ll ccccccc c ccccccc@{}}
\toprule
 & & \multicolumn{7}{c}{8-action budget} & & \multicolumn{7}{c}{16-action budget} \\
\cmidrule(lr){3-9}\cmidrule(l){11-17}
Model & Arm & mean3 & pers. & ans. & med. & final & calls & \ii &  & mean3 & pers. & ans. & med. & final & calls & \ii \\
\midrule
\qtwo & none & .208 & .043 & 156 & 2 & .00 & 3.9 & $-.35$ &  & .219 & .043 & 158 & 2 & .01 & 7.1 & $-.35$ \\
 & budget & .300 & .090 & 175 & 8 & .75 & 6.8 & $-.02$ &  & .312 & .093 & 232 & 16 & .75 & 13.4 & $-.03$ \\
 & rule & .246 & .100 & 262 & 2 & .00 & 2.8 & .09 &  & .249 & .093 & 260 & 2 & .00 & 2.8 & .10 \\
 & combo & .357 & .163 & 282 & 6 & .00 & 4.7 & .45 &  & .357 & .130 & 267 & 6 & .00 & 4.7 & .44 \\
\midrule
\llama & none & .241 & .000 & 4 & 4.5 & .25 & 7.9 & .00 &  & .298 & .010 & 16 & 13 & .06 & 15.6 & .00 \\
 & budget & .283 & .057 & 215 & 8 & .79 & 7.0 & .03 &  & .299 & .050 & 183 & 16 & .70 & 14.4 & .00 \\
 & rule & .314 & .143 & 285 & 6 & .01 & 5.1 & .47 &  & .377 & .170 & 296 & 6 & .00 & 5.2 & .47 \\
 & combo & .309 & .177 & 296 & 6 & .03 & 5.0 & .48 &  & .349 & .177 & 294 & 6 & .00 & 5.0 & .49 \\
\midrule
\qthree & none & .382 & .007 & 8 & 3 & .12 & 7.9 & $-.01$ &  & .393 & .013 & 13 & 5 & .08 & 15.5 & $-.01$ \\
 & budget & .426 & .203 & 235 & 8 & .51 & 5.8 & $-.03$ &  & .433 & .183 & 198 & 16 & .56 & 12.0 & $-.05$ \\
 & rule & .433 & .153 & 284 & 6 & .00 & 5.1 & .46 &  & .442 & .160 & 286 & 6 & .00 & 5.5 & .46 \\
 & combo & .447 & .270 & 300 & 6 & .01 & 4.3 & .42 &  & .436 & .220 & 300 & 6 & .00 & 4.4 & .42 \\
\midrule
\qthirty & none & .438 & .027 & 34 & 6 & .18 & 7.6 & .01 &  & -- &  &  &  &  &  &  \\
 & budget & .551 & .330 & 298 & 6 & .04 & 4.6 & .29 &  & .551 & .347 & 300 & 7 & .01 & 6.7 & .10 \\
 & rule & .526 & .267 & 294 & 6 & .00 & 5.0 & .48 &  & .543 & .290 & 294 & 6 & .00 & 5.0 & .48 \\
 & combo & .533 & .310 & 298 & 6 & .00 & 4.2 & .51 &  & .546 & .323 & 298 & 6 & .00 & 4.3 & .46 \\
\bottomrule
\end{tabular}
\caption{Budget scaling on HotpotQA (test300, raw analysis): each arm with the 8- and the 16-action budget. The 16-action runs cover three regimes (persistent, recover\_after\_2, clean), so mean3 averages these three at both budgets in place of mean6. On the persistent regime: success (pers.), number of the 300 questions on which the agent answers rather than exhausting its budget or failing to produce a well-formed action (ans.), median action at which it answers, among those (med., 1-based), share of those answers given at the final action (final), and mean tool calls (search and lookup) per question (calls). \ii is computed as in Table~\ref{tab:app-hotpot} but over the persistent and recover\_after\_2 regimes only, with the final action of the respective budget excluded; its 8-action values therefore differ from Table~\ref{tab:app-hotpot}. $n=300$ per cell; \qthirty without intervention was not run with 16 actions.}
\label{tab:app-budget16}
\end{table*}

\subsection{Fixed step caps, the best simple policy, and whether stopping pays}\label{app:cap}
These analyses are exploratory. They evaluate stopping policies offline on the unaided runs: a policy that stops at a given decision point is credited with the answer-now branch replayed from that point (the same forced-finish instruction the rule uses; Appendix~\ref{app:prompts}), or with the unaided run's own outcome if the agent had already answered. Answer-now branches exist for four regimes, so these comparisons average those (mean4).

\paragraph{Continue or answer now.} Would stopping on the agent's own judgments have paid? We take the first decision point in each unaided run where the agent's replayed judgments reach five consecutive USELESS (the four regimes with answer-now branches). There we compare the success of answering right then (the answer-now branch) with that of what the agent actually did; it almost always continued. Answering then succeeds more often for all three models: \qtwo 0.171 against 0.065 (difference $+0.106$ [$+0.065, +0.145$]; $n=461$; answering better on 67 and continuing on 18 decisions); \llama 0.159 against 0.005 (difference $+0.154$ [$+0.119, +0.191$]; $n=610$; answering better on 96 and continuing on 2 decisions); \qthree 0.149 against 0.013 (difference $+0.136$ [$+0.104, +0.169$]; $n=544$; answering better on 77 and continuing on 3 decisions). Intervals are 95\% question-bootstrap. Continuing spent 2.5--2.6 further tool calls on average. The comparison is exploratory and specific to sources that, once failed five times, do not recover.

\paragraph{Best simple policy.} Table~\ref{tab:norm} evaluates offline, on the unaided runs, every policy that answers after the $c$-th observation (cap $c$), after $k$ consecutive useless side-channel judgments (own $k$; judgments replayed on the unaided trajectories), or whichever comes first, using the answer-now branch at the decision where it fires (the four regimes with answer-now branches: clean, persistent, plausible and recover\_after\_2). Utility is success minus $\lambda$ times tool calls; the best policy is chosen on the same data. Regret is the best policy's utility minus the arm's (the arms' own runs; negative means the arm is better). The offline own-5 policy reproduces the rule arm (success .378 vs.\ .372, .271 vs.\ .271, .216 vs.\ .224 for \qthree, \llama, \qtwo).

\begin{table*}[tp]
\centering\footnotesize
\begin{adjustbox}{max width=\linewidth}\begin{tabular}{@{}llllccccccc@{}}
\toprule
& & & & \multicolumn{6}{c}{regret of} \\
\cmidrule(l){5-10}
Model & $\lambda$ & best policy & utility & none & permit & budget & cost & rule & combo \\
\midrule
\qtwo & 0.0 & own 7 & 0.232 & +0.057 & +0.028 & $-$0.039 & $-$0.042 & +0.008 & $-$0.097 \\
 & 0.02 & cap 3 & 0.174 & +0.082 & +0.041 & +0.005 & +0.003 & +0.014 & $-$0.073 \\
 & 0.05 & cap 3 & 0.108 & +0.140 & +0.080 & +0.092 & +0.090 & +0.043 & $-$0.017 \\
\llama & 0.0 & own 3 or cap 5 & 0.322 & +0.129 & +0.120 & +0.075 & +0.074 & +0.051 & +0.050 \\
 & 0.02 & own 3 or cap 5 & 0.252 & +0.188 & +0.153 & +0.114 & +0.113 & +0.085 & +0.075 \\
 & 0.05 & own 3 or cap 4 & 0.148 & +0.277 & +0.205 & +0.173 & +0.171 & +0.138 & +0.114 \\
\qthree & 0.0 & cap 7 & 0.388 & +0.071 & +0.027 & +0.011 & +0.023 & +0.015 & $-$0.013 \\
 & 0.02 & own 3 or cap 7 & 0.321 & +0.112 & +0.018 & +0.025 & +0.040 & +0.034 & $-$0.009 \\
 & 0.05 & own 3 or cap 5 & 0.227 & +0.180 & +0.010 & +0.052 & +0.071 & +0.067 & +0.004 \\
\bottomrule
\end{tabular}\end{adjustbox}
\caption{Distance from the best simple stopping policy (exploratory; mean over clean, persistent, plausible, recover\_after\_2). cost: the stated-cost arm, whose prompt gives the scoring with $\lambda=0.05$.}
\label{tab:norm}
\end{table*}

\paragraph{Calls under a doubled budget.} Table~\ref{tab:cost} gives tool calls per trajectory at 8 and 16 actions: the budget cue's calls grow with the budget, whereas the rule's and the combination's barely change.

\begin{table}[tp]
\centering\footnotesize\setlength{\tabcolsep}{4pt}
\begin{tabular}{@{}l ccc@{}}
\toprule
Model & budget & rule & combo \\
\midrule
\qtwo & 5.09 $\to$ 8.69 & 2.88 $\to$ 2.91 & 4.06 $\to$ 4.10 \\
\llama & 5.45 $\to$ 8.81 & 5.11 $\to$ 5.39 & 4.74 $\to$ 4.99 \\
\qthree & 4.11 $\to$ 6.62 & 4.10 $\to$ 4.27 & 3.54 $\to$ 3.62 \\
\bottomrule
\end{tabular}
\caption{Tool calls per trajectory with an 8- and a 16-action budget (persistent, recover\_after\_2 and clean, the regimes run at both budgets; exploratory).}
\label{tab:cost}
\end{table}

\subsection{The rule on stated judgments, and \haiku\ with the rule or cost stated}\label{app:add19}
Amendment 19 replaces the enforced rule's side-channel question with a reading of the judgment the agent already writes in its reasoning; amendment 20 runs the stated-rule and call-cost conditions on \haiku. Both were frozen before their test runs (Table~\ref{tab:prereg}).

\paragraph{Reading the stated judgment.} After every observation the agent writes its usual thought and action. Before the action is executed, a keyword reader (no model call) classifies the first two sentences of the thought as useless (the agent says the result does not help), useful (it reports something usable) or no judgment, and keeps the run of stated-useless judgments: useless adds one, useful resets it, no judgment leaves it unchanged. When the run reaches five and the proposed action is a search or lookup, that turn is discarded and the agent is asked again with the enforced rule's finish instruction, so the rule fires at the same step as with the side channel. The reader was written on development data only: on the 1,675 adjudicated stated judgments of \qthree's development trajectories its useless readings have precision 0.971 and recall 0.913. The stated-judgment \emph{combination} adds the budget statement. Agents do not always write a judgment: the reader finds none at 0.25, 0.15 and 0.22 of decision points for \llama, \qthree and \qthirty, but at 0.74 for \qtwo, whose thoughts are mostly empty (0.39 when told its budget). We count every call. The side-channel rule makes one judgment call after each observation, and the stated-judgment rule one extra generation each time it fires. Utility is success minus 0.05 per tool or extra call, which is the price in the call-cost condition. Table~\ref{tab:a19} gives success, calls and utility for every condition, and Table~\ref{tab:a19tests} the pre-registered tests.

\begin{table*}[tp]
\centering\footnotesize\setlength{\tabcolsep}{5pt}
\begin{tabular}{@{}ll cccc@{}}
\toprule
Model & Condition & mean6 & tool calls & extra calls & utility ($\lambda=0.05$) \\
\midrule
\qtwo & none & 0.192 & 3.63 & 0.00 & $+0.010$ \\
 & budget & 0.328 & 4.90 & 0.00 & $+0.083$ \\
 & enforced rule & 0.237 & 2.94 & 2.86 & $-0.053$ \\
 & combo & 0.375 & 4.03 & 4.00 & $-0.027$ \\
 & stated-judgment rule & 0.209 & 3.46 & 0.08 & $+0.033$ \\
 & budget + stated-judgment rule & 0.343 & 4.42 & 0.24 & $+0.110$ \\
\addlinespace[2pt]
\llama & none & 0.263 & 6.13 & 0.00 & $-0.043$ \\
 & budget & 0.312 & 5.38 & 0.00 & $+0.043$ \\
 & enforced rule & 0.328 & 5.31 & 5.09 & $-0.192$ \\
 & combo & 0.337 & 4.88 & 4.85 & $-0.150$ \\
 & stated-judgment rule & 0.328 & 5.33 & 0.35 & $+0.044$ \\
 & budget + stated-judgment rule & 0.346 & 4.90 & 0.27 & $+0.088$ \\
\addlinespace[2pt]
\qthree & none & 0.448 & 4.79 & 0.00 & $+0.208$ \\
 & budget & 0.464 & 3.85 & 0.00 & $+0.272$ \\
 & enforced rule & 0.474 & 4.09 & 4.01 & $+0.069$ \\
 & combo & 0.478 & 3.48 & 3.47 & $+0.130$ \\
 & stated-judgment rule & 0.484 & 4.07 & 0.31 & $+0.265$ \\
 & budget + stated-judgment rule & 0.478 & 3.50 & 0.18 & $+0.294$ \\
\addlinespace[2pt]
\qthirty & none & 0.509 & 4.76 & 0.00 & $+0.271$ \\
 & budget & 0.581 & 3.64 & 0.00 & $+0.399$ \\
 & enforced rule & 0.572 & 4.10 & 4.05 & $+0.165$ \\
 & combo & 0.578 & 3.54 & 3.54 & $+0.224$ \\
 & stated-judgment rule & 0.573 & 4.12 & 0.26 & $+0.355$ \\
 & budget + stated-judgment rule & 0.580 & 3.55 & 0.06 & $+0.399$ \\
\bottomrule
\end{tabular}
\caption{Stopping on the judgment stated in the agent's reasoning (amendment 19; test300, raw, per question averaged over the six mean6 regimes). Extra calls: side-channel judgments for the enforced rule and the combination, discarded turns for the stated-judgment conditions.}
\label{tab:a19}
\end{table*}

\begin{table*}[tp]
\centering\footnotesize\setlength{\tabcolsep}{4pt}
\begin{tabular}{@{}l cccc c@{}}
\toprule
& E1 & E2 & E3 & E4 & \\
Model & \makecell{success\\vs none} & \makecell{success vs\\enforced rule} & \makecell{utility vs\\enforced rule} & \makecell{combination\\utility vs budget} & \makecell{fires\\(pers.)} \\
\midrule
\qtwo & \makecell{$+0.018^{*}$\\[-1pt]{\scriptsize[$+0.011, +0.024$]}} & \makecell{$-0.027$\\[-1pt]{\scriptsize[$-0.041, -0.012$]}} & \makecell{$+0.086^{*}$\\[-1pt]{\scriptsize[$+0.070, +0.102$]}} & \makecell{$+0.027^{*}$\\[-1pt]{\scriptsize[$+0.018, +0.037$]}} & 0.05 \\
\llama & \makecell{$+0.064^{*}$\\[-1pt]{\scriptsize[$+0.049, +0.079$]}} & \makecell{$+0.000$\\[-1pt]{\scriptsize[$-0.017, +0.018$]}} & \makecell{$+0.236^{*}$\\[-1pt]{\scriptsize[$+0.216, +0.257$]}} & \makecell{$+0.045^{*}$\\[-1pt]{\scriptsize[$+0.032, +0.059$]}} & 0.91 \\
\qthree & \makecell{$+0.036^{*}$\\[-1pt]{\scriptsize[$+0.022, +0.051$]}} & \makecell{$+0.010$\\[-1pt]{\scriptsize[$-0.002, +0.021$]}} & \makecell{$+0.196^{*}$\\[-1pt]{\scriptsize[$+0.182, +0.210$]}} & \makecell{$+0.022^{*}$\\[-1pt]{\scriptsize[$+0.009, +0.034$]}} & 0.97 \\
\qthirty & \makecell{$+0.064^{*}$\\[-1pt]{\scriptsize[$+0.047, +0.081$]}} & \makecell{$+0.001$\\[-1pt]{\scriptsize[$-0.014, +0.016$]}} & \makecell{$+0.190^{*}$\\[-1pt]{\scriptsize[$+0.173, +0.205$]}} & \makecell{$0.000$\\[-1pt]{\scriptsize[$-0.007, +0.006$]}} & 0.89 \\
\bottomrule
\end{tabular}
\caption{Pre-registered tests of amendment 19 for the stated-judgment rule (E1--E3) and its combination with the budget (E4); mean6 or utility at $\lambda=0.05$; question-paired 95\% intervals; $^{*}$ one-sided Holm-corrected $p<0.05$; E2 is a non-inferiority test with margin $-0.05$, passed by all four models. By the pre-registered applicability rule, a model whose stated-judgment rule fires on fewer than 20\% of persistently failing questions does not count as support.}
\label{tab:a19tests}
\end{table*}

\paragraph{\haiku\ with the rule or the cost stated (amendment 20).} The first 200 test300 questions, six regimes, compared with \haiku's unaided and enforced-rule runs on the same questions (unaided mean6 0.518, enforced rule 0.611). Under these prompts \haiku\ often wrote its native tool-call markup instead of an action line, or answered in prose without the finish action; the runs therefore read the first tool call as the action and send one fixed format reminder when a turn has no readable action. Both fixes were set before the runs. The earlier unaided and enforced-rule runs lacked them and had 2--3\% format failures and a further 2\% truncated generations on these questions, which favors the new conditions, most of all the call-cost condition with almost no format failures. With the price stated, utility at $\lambda=0.05$ exceeds the unaided agent's by $+0.154$ and the enforced rule's by $+0.213$ (the latter counting its judgment calls). Table~\ref{tab:a20} gives the results. The stated-rule verdict uses the rule of amendment 11 unchanged: \emph{partial}.

\begin{table*}[tp]
\centering\footnotesize\setlength{\tabcolsep}{4pt}
\begin{tabular}{@{}l c ccc cc@{}}
\toprule
\haiku & mean6 & $-$ unaided & $-$ enforced rule & \makecell{design-label\\contrast} & \makecell{answered\\(pers.)} & \makecell{median\\action} \\
\midrule
stated rule & 0.561 & \makecell{$+0.043$\\[-1pt]{\scriptsize[$+0.007, +0.077$]}} & \makecell{$-0.050$\\[-1pt]{\scriptsize[$-0.086, -0.017$]}} & \makecell{$+0.05$\\[-1pt]{\scriptsize[$+0.02, +0.09$]}} & 0.565 & 7 \\
call cost & 0.663 & \makecell{$+0.145$\\[-1pt]{\scriptsize[$+0.113, +0.182$]}} & \makecell{$+0.053$\\[-1pt]{\scriptsize[$+0.023, +0.082$]}} & \makecell{$+0.01$\\[-1pt]{\scriptsize[$-0.02, +0.05$]}} & 0.985 & 7 \\
\bottomrule
\end{tabular}
\caption{\haiku\ with the stopping rule or the call cost stated (amendment 20; first 200 test300 questions, raw; question-paired 95\% intervals).}
\label{tab:a20}
\end{table*}

\subsection{Fresh questions, stated judgments, reasoning mode and Search-R1}\label{app:add16}
Amendments 15--17 were frozen before any of their runs (Table~\ref{tab:prereg}).

\paragraph{Fresh questions (amendment 16a).} Before any run on them, we sampled 300 new questions from the HotpotQA distractor development set with a new random seed, disjoint from every question used before (pilot, development and test; the list is in the code release). We ran them with the settings of test300. Table~\ref{tab:fresh} gives mean6 and the own-judgment contrast $\Delta$ (judgments replayed for none and budget, recorded for the rule) with the design-label contrast; $\Delta$ was the primary endpoint.

\begin{table*}[tp]
\centering\footnotesize\setlength{\tabcolsep}{3pt}
\begin{adjustbox}{max width=\linewidth}\begin{tabular}{@{}l ccc ccc c@{}}
\toprule
& \multicolumn{3}{c}{mean6} & \multicolumn{3}{c}{$\Delta$ own judgments (second line: design labels)} & answered after \\
Model & none & budget & rule & none & budget & rule & 5 useless (none) \\
\midrule
\qtwo & 0.175 & 0.315 & 0.205 & $-0.09$ [$-0.16, -0.03$] & $-0.17$ [$-0.24, -0.11$] & $+0.31$ [$+0.22, +0.38$] & 3/125 \\
 &  &  &  & {\scriptsize $-0.02$ [$-0.04, 0.00$]} & {\scriptsize $-0.06$ [$-0.08, -0.04$]} & {\scriptsize $+0.11$ [$+0.07, +0.14$]} &  \\
\llama & 0.234 & 0.281 & 0.294 & $-0.03$ [$-0.05, -0.02$] & $-0.14$ [$-0.16, -0.11$] & $+0.35$ [$+0.33, +0.37$] & 2/283 \\
 &  &  &  & {\scriptsize $-0.01$ [$-0.02, 0.00$]} & {\scriptsize $-0.03$ [$-0.05, -0.01$]} & {\scriptsize $+0.17$ [$+0.14, +0.20$]} &  \\
\qthree & 0.422 & 0.453 & 0.458 & $-0.05$ [$-0.07, -0.03$] & $-0.19$ [$-0.24, -0.14$] & $+0.34$ [$+0.31, +0.38$] & 0/289 \\
 &  &  &  & {\scriptsize $-0.01$ [$-0.02, 0.00$]} & {\scriptsize $-0.10$ [$-0.14, -0.06$]} & {\scriptsize $+0.16$ [$+0.12, +0.20$]} &  \\
\qthirty & 0.477 & 0.528 & 0.534 & $-0.06$ [$-0.09, -0.04$] & $-0.05$ [$-0.10, +0.01$] & $+0.30$ [$+0.26, +0.34$] & 19/284 \\
 &  &  &  & {\scriptsize $-0.02$ [$-0.04, 0.00$]} & {\scriptsize $-0.05$ [$-0.11, 0.00$]} & {\scriptsize $+0.19$ [$+0.14, +0.22$]} &  \\
\bottomrule
\end{tabular}\end{adjustbox}
\caption{Confirmatory replication on fresh300 (amendment 16a), raw analysis, 95\% intervals.}
\label{tab:fresh}
\end{table*}

\paragraph{Stated and side-channel judgments (amendment 16b).} A local \qthirty annotator (non-thinking, temperature 0, the rubric of the LLM annotators) labeled the 327 adjudicated stated-judgment items with five-class $\kappa$ 0.842 [0.766, 0.910] and binary (useless or not) $\kappa$ 0.874 [0.798, 0.939]. We took 160 decision points per model from the unaided test300 runs (80 after a failed and 80 after a real observation). Where the agent states a judgment in its next thought, that judgment agrees with the side-channel judgment replayed at the same point as follows: \qtwo 0.87 ($\kappa$ 0.60; 99 of 160 explicit); \llama 0.94 ($\kappa$ 0.84; 155 of 160 explicit); \qthree 0.94 ($\kappa$ 0.84; 156 of 160 explicit); \qthirty 0.96 ($\kappa$ 0.89; 157 of 160 explicit).

\paragraph{Stopping on stated judgments (amendment 18).} The same annotator labeled the judgment stated in the next thought at every decision point of the unaided test300 runs (failure regimes; decision index $\le 6$, persistent $\le 7$). Stated useful or partly useful judgments reset the run; points without an explicit judgment make the run unknown until the next useful one, and decisions with an unknown run are left out. Table~\ref{tab:stated} gives $\Delta$ and the dissociation computed from these stated judgments next to the side-channel versions.

\begin{table*}[tp]
\centering\footnotesize\setlength{\tabcolsep}{7pt}
\begin{adjustbox}{max width=\linewidth}\begin{tabular}{@{}l c c c c c@{}}
\toprule
Model & explicit & $\Delta$ stated & $\Delta$ side channel & $n$ (all useless / ref.) & answered after 5 \\
\midrule
\qtwo & 65\% & $-0.08$ [$-0.15, -0.03$] & $-0.11$ [$-0.17, -0.05$] & 304 / 136 & 0/15 \\
\llama & 96\% & $-0.01$ [$-0.02, -0.01$] & $-0.02$ [$-0.03, -0.02$] & 1911 / 1029 & 1/266 \\
\qthree & 99\% & $-0.04$ [$-0.06, -0.01$] & $-0.06$ [$-0.10, -0.04$] & 2072 / 380 & 1/287 \\
\qthirty & 98\% & $-0.02$ [$-0.05, 0.00$] & $-0.07$ [$-0.10, -0.04$] & 1787 / 418 & 16/252 \\
\bottomrule
\end{tabular}\end{adjustbox}
\caption{Time-matched contrast and dissociation from the judgments agents state in their own reasoning (amendment 18), unaided agents, test300. explicit: share of decision points with an explicit stated judgment.}
\label{tab:stated}
\vspace{8pt}
\centering\footnotesize\setlength{\tabcolsep}{7pt}
\begin{adjustbox}{max width=\linewidth}\begin{tabular}{@{}l cccc cc@{}}
\toprule
\qthirty & mean4 & ans. & med. & prem. & $\Delta$ own & $\Delta$ design \\
\midrule
none & 0.464 & 0.11 & 6 & 0.02 & $-0.07$ [$-0.10, -0.04$] & $-0.02$ [$-0.04, 0.00$] \\
none, thinking & 0.540 & 0.97 & 3 & 0.47 & $+0.01$ [$-0.18, +0.17$] & $0.00$ [$-0.14, +0.12$] \\
budget & 0.576 & 0.99 & 6 & 0.11 & $0.00$ [$-0.06, +0.05$] & $-0.03$ [$-0.09, +0.02$] \\
budget, thinking & 0.525 & 1.00 & 3 & 0.77 & $+0.25$ [$-0.01, +0.50$] & $+0.32$ [$+0.10, +0.52$] \\
\bottomrule
\end{tabular}\end{adjustbox}
\caption{\qthirty with and without thinking (test300, four regimes). ans.: share answered on a persistently failing source; med.: median answering action there; prem.: share of recover\_after\_2 questions answered before any real result.}
\label{tab:think}
\end{table*}

\paragraph{Reasoning mode (amendment 15).} \qthirty ran with thinking enabled (Qwen3's recommended thinking sampling, 12,288-token context; up to 8,192 new tokens per step instead of the pre-registered 2,048, deviation 6) on persistent, late\_onset\_from\_3, recover\_after\_2 and clean; the stated-cost arm was not run (deviation 4). Thinking minus non-thinking, mean over the four regimes: none $+0.076$ [$+0.038, +0.112$], budget $-0.051$ [$-0.083, -0.018$] (Table~\ref{tab:think}).

\paragraph{Search-R1 (amendment 17).} We ran the released Search-R1 checkpoint (Qwen2.5-7B-Instruct, PPO, trained on NQ and HotpotQA training questions) with its own prompt, \texttt{<search>}/\allowbreak\texttt{<information>}/\allowbreak\texttt{<answer>} tags and one document per search; knowledge bases, failure regimes, the 8-action budget and test300 as in the main runs; greedy decoding. The side-channel question was asked after every observation on a copy of the conversation. Its judgments call 100\% of failed and 89\% of real observations useless (pre-registered validity: at most 50\% of real ones; the other open models' rule-arm logs on the same six regimes: 35--73\%), so its own-judgment contrast ($-0.26$ [$-0.58, -0.06$]) is not interpreted. Unaided, it answers on 53\% of persistently failing questions (median action 3); design-label contrast $-0.03$ [$-0.07, +0.01$]. The rule fired 498 times over the six mean6 regimes, and the agent answered after 424 of them; on a persistently failing source it ends without a well-formed answer on 12\% of questions (none: 3\%). Rule minus none, mean6: $+0.024$ [$+0.009, +0.041$].

\subsection{Backup tool and harder failures}\label{app:add13}
Amendment 13 was frozen before its runs; all runs are local.

\paragraph{Two sources.} The prompt adds \texttt{backup\_\allowbreak search[query]}: ``Search a second, independent copy of the knowledge base. It is slower and more expensive than search: prefer search, and use backup\_search when search does not give you useful results.'' The backup searches the intact knowledge base and never fails; the failure schedules apply to the primary search. The switch rule disables the primary search after five consecutive side-channel ``useless'' judgments of its results and appends ``The search tool is no longer available. Use backup\_search[query] or finish[answer].'' Table~\ref{tab:ts} reports success, the timing of the first switch on a persistently failing primary, and the own-judgment time-matched contrast for leaving the primary (switching or answering; judgments replayed for the unaided and budget arms, logged for the switch rule). By the pre-registered interpretation, a model ``switches'' if its share switched by the fourth action is at least 0.5, and otherwise ``does not abandon'' if the contrast's interval includes or lies below 0. \qtwo and \llama therefore do not abandon, and \qthree and \qthirty switch. Switch rule minus none, mean over the six regimes: \qtwo $+0.011$ [$-0.005, +0.026$]; \llama $+0.054$ [$+0.037, +0.071$]; \qthree $+0.016$ [$+0.005, +0.026$]; \qthirty $+0.007$ [$-0.008, +0.024$].

\begin{table*}[tp]
\centering\footnotesize\setlength{\tabcolsep}{3.5pt}
\begin{adjustbox}{max width=\linewidth}\begin{tabular}{@{}llccccccc c@{}}
\toprule
& & \multicolumn{3}{c}{success} & \multicolumn{4}{c}{persistently failing primary} & \\
\cmidrule(lr){3-5}\cmidrule(lr){6-9}
Model & Arm & clean & persistent & mean6 & switched by 4 & ever switched & median first & answered first & $\Delta$ (leave) \\
\midrule
\qtwo & none & 0.480 & 0.230 & 0.383 & 0.49 & 0.62 & 2 & 0.02 & $-0.54$ \\
 & budget & 0.490 & 0.240 & 0.400 & 0.50 & 0.61 & 3 & 0.11 & $-0.38$ \\
 & switch rule & 0.470 & 0.293 & 0.393 & 0.51 & 0.97 & 4 & 0.02 & $-0.29$ \\
\llama & none & 0.497 & 0.050 & 0.334 & 0.17 & 0.33 & 4 & 0.01 & $-0.06$ \\
 & budget & 0.477 & 0.130 & 0.356 & 0.15 & 0.54 & 6 & 0.31 & $-0.04$ \\
 & switch rule & 0.540 & 0.223 & 0.388 & 0.19 & 0.91 & 6 & 0.06 & $+0.26$ \\
\qthree & none & 0.593 & 0.367 & 0.529 & 0.73 & 0.79 & 2 & 0.01 & $-0.09$ \\
 & budget & 0.607 & 0.347 & 0.523 & 0.31 & 0.37 & 3 & 0.44 & $-0.24$ \\
 & switch rule & 0.600 & 0.477 & 0.544 & 0.73 & 0.99 & 2.5 & 0.01 & $+0.30$ \\
\qthirty & none & 0.677 & 0.597 & 0.639 & 0.95 & 0.99 & 2 & 0.01 & $+0.02$ \\
 & budget & 0.710 & 0.620 & 0.672 & 0.87 & 0.96 & 3 & 0.04 & $+0.14$ \\
 & switch rule & 0.673 & 0.603 & 0.646 & 0.95 & 0.99 & 2 & 0.01 & $+0.03$ \\
\bottomrule
\end{tabular}\end{adjustbox}
\caption{Two-source setting, HotpotQA test300, raw analysis. On a failing primary source: ``switched by 4'', share that switched to the backup by the fourth action; ``median first'', median action of the first switch, among questions where it switched; ``answered first'', share that answered before any switch; $\Delta$ (leave), own-judgment contrast for leaving the primary source by either exit.}
\label{tab:ts}
\end{table*}

\paragraph{Answerless source and long recovery.} In the answerless regime every page lacks the question's supporting-fact sentences. The unaided agent's replayed judgments call these pages useless 97\% (\qtwo), 81\% (\llama), 86\% (\qthree) and 76\% (\qthirty) of the time. After five consecutive useless judgments, it answers on 41 of 202, 9 of 166, 26 of 170 and 26 of 122 questions, respectively. Table~\ref{tab:al} gives success and the rule's contrast with the unaided agent; the pre-registered prediction that the rule loses in recover\_after\_5 holds for the Qwen3 models and not for \llama\ and \qtwo, and non-inferiority in recover\_after\_4 holds for three models and narrowly fails for \qthree\ (lower bound $-0.050$).

\begin{table*}[tp]
\centering\footnotesize\setlength{\tabcolsep}{3pt}
\begin{adjustbox}{max width=\linewidth}\begin{tabular}{@{}llccc@{}}
\toprule
Model & Arm & answerless & rec.\ after 4 & rec.\ after 5 \\
\midrule
\qtwo & none & 0.137 & 0.117 & 0.087 \\
 & budget & 0.217 & 0.317 & 0.200 \\
 & rule & 0.183 & 0.167 & 0.093 \\
 & combo & 0.273 & 0.253 & 0.163 \\
\llama & none & 0.107 & 0.223 & 0.117 \\
 & budget & 0.173 & 0.287 & 0.187 \\
 & rule & 0.153 & 0.210 & 0.153 \\
 & combo & 0.210 & 0.270 & 0.167 \\
\qthree & none & 0.173 & 0.473 & 0.410 \\
 & budget & 0.260 & 0.447 & 0.397 \\
 & rule & 0.203 & 0.453 & 0.197 \\
 & combo & 0.273 & 0.420 & 0.277 \\
\qthirty & none & 0.227 & 0.517 & 0.427 \\
 & budget & 0.353 & 0.513 & 0.403 \\
 & rule & 0.253 & 0.533 & 0.290 \\
 & combo & 0.347 & 0.473 & 0.317 \\
\midrule
\multicolumn{5}{@{}l}{rule $-$ none} \\
\qtwo & & \begin{tabular}[c]{@{}c@{}}$+0.047$\\[-2pt]{\scriptsize[$+0.013, +0.080$]}\end{tabular} & \begin{tabular}[c]{@{}c@{}}$+0.050$\\[-2pt]{\scriptsize[$+0.013, +0.087$]}\end{tabular} & \begin{tabular}[c]{@{}c@{}}$+0.007$\\[-2pt]{\scriptsize[$-0.020, +0.033$]}\end{tabular} \\
\llama & & \begin{tabular}[c]{@{}c@{}}$+0.047$\\[-2pt]{\scriptsize[$+0.003, +0.087$]}\end{tabular} & \begin{tabular}[c]{@{}c@{}}$-0.013$\\[-2pt]{\scriptsize[$-0.047, +0.023$]}\end{tabular} & \begin{tabular}[c]{@{}c@{}}$+0.037$\\[-2pt]{\scriptsize[$-0.013, +0.090$]}\end{tabular} \\
\qthree & & \begin{tabular}[c]{@{}c@{}}$+0.030$\\[-2pt]{\scriptsize[$+0.000, +0.060$]}\end{tabular} & \begin{tabular}[c]{@{}c@{}}$-0.020$\\[-2pt]{\scriptsize[$-0.050, +0.013$]}\end{tabular} & \begin{tabular}[c]{@{}c@{}}$-0.213$\\[-2pt]{\scriptsize[$-0.273, -0.153$]}\end{tabular} \\
\qthirty & & \begin{tabular}[c]{@{}c@{}}$+0.027$\\[-2pt]{\scriptsize[$-0.010, +0.063$]}\end{tabular} & \begin{tabular}[c]{@{}c@{}}$+0.017$\\[-2pt]{\scriptsize[$-0.030, +0.060$]}\end{tabular} & \begin{tabular}[c]{@{}c@{}}$-0.137$\\[-2pt]{\scriptsize[$-0.200, -0.077$]}\end{tabular} \\
\bottomrule
\end{tabular}\end{adjustbox}
\caption{Answerless source and long recovery, raw analysis.}
\label{tab:al}
\end{table*}

\subsection{Second and third runs}\label{app:rep}
Amendment 10 repeated the none, budget, rule and combination arms on the three 7--8B models with new sampling seeds (Qwen3) and a new draw of the failed observations (all models), keeping questions, regimes and prompts fixed; it was frozen before the replicate runs. Table~\ref{tab:rep} gives the replicate's results and pre-registered tests (raw). Rp1 (rule $-$ none $>0$, Holm) passes for all three models. Rp2 (rule index lower bound $>+0.30$) fails for \qtwo, whose index was $0.18$ in the original run as well, while the unaided index stays at or below $+0.05$. Rp3 (budget arm answers at the deadline, index $\le+0.10$) passes for all three. Rp4 (combination non-inferior to both components, margin $-0.05$) passes five of six tests and fails against the rule for \qthree (lower bound $-0.054$), but passes all six after answer repair ($+0.008$ [$-0.020, +0.036$] for \qthree). By amendment 10's interpretation rule, these two conclusions are not stable in the replicate run. All other outcomes are unchanged by the repair. Replicate minus original mean6 is within $\pm0.023$ for every model and arm.

\begin{table*}[tp]
\centering\footnotesize\setlength{\tabcolsep}{4pt}
\begin{adjustbox}{max width=\linewidth}\begin{tabular}{@{}lcccc ccccc@{}}
\toprule
& \multicolumn{4}{c}{mean6 (replicate)} & rule $-$ none & rule \ii\ [LB] & none \ii & budget median & rule design $\Delta^e$ \\
Model & none & budget & rule & combo & (Rp1) & (Rp2) & (Rp2) & action (Rp3) & \\
\midrule
\qtwo & 0.209 & 0.314 & 0.257 & 0.366 & $+0.048$ [$+0.033, +0.063$] & 0.19 [0.14] & $-0.31$ & 8 & 0.06 \\
\llama & 0.264 & 0.307 & 0.325 & 0.332 & $+0.061$ [$+0.043, +0.078$] & 0.46 [0.44] & $0.00$ & 8 & 0.17 \\
\qthree & 0.455 & 0.456 & 0.497 & 0.473 & $+0.042$ [$+0.029, +0.056$] & 0.46 [0.44] & $0.00$ & 8 & 0.15 \\
\bottomrule
\end{tabular}\end{adjustbox}

\vspace{6pt}
\begin{adjustbox}{max width=\linewidth}\begin{tabular}{@{}lcccc@{}}
\toprule
& \multicolumn{4}{c}{replicate $-$ original mean6 (paired by question)} \\
Model & none & budget & rule & combo \\
\midrule
\qtwo & $+0.018$ [$-0.006, +0.041$] & $-0.014$ [$-0.036, +0.008$] & $+0.021$ [$-0.003, +0.044$] & $-0.009$ [$-0.031, +0.013$] \\
\llama & $+0.001$ [$-0.017, +0.019$] & $-0.005$ [$-0.024, +0.014$] & $-0.003$ [$-0.021, +0.014$] & $-0.004$ [$-0.024, +0.017$] \\
\qthree & $+0.007$ [$-0.013, +0.026$] & $-0.009$ [$-0.036, +0.017$] & $+0.023$ [$+0.003, +0.043$] & $-0.004$ [$-0.031, +0.021$] \\
\bottomrule
\end{tabular}\end{adjustbox}
\caption{Pre-registered replicate (amendment 10), raw analysis, HotpotQA test300. Top: replicate results and tests; bottom: replicate minus original. $^e$Design-label contrast, exploratory.}
\label{tab:rep}
\end{table*}

\paragraph{Third run.} Table~\ref{tab:rep3} repeats amendment 10's tests on a third run and pools the rule's gain over all three runs, with a bootstrap over runs and questions.

\begin{table*}[tp]
\centering\footnotesize\setlength{\tabcolsep}{3pt}
\begin{adjustbox}{max width=\linewidth}\begin{tabular}{@{}lcccccc@{}}
\toprule
Model & rule $-$ none (Rp1) & rule \ii\ [lo] & none \ii & budget median / \ii\ (Rp3) & combo LB vs rule / budget (Rp4) & pooled rule $-$ none \\
\midrule
\qtwo & \begin{tabular}[c]{@{}c@{}}$+0.047$\\[-2pt]{\scriptsize[$+0.032, +0.063$]}\end{tabular} & 0.17 [0.12] & $-$0.34 & 8 / $-$0.02 & +0.082 / +0.024 & \begin{tabular}[c]{@{}c@{}}$+0.047$\\[-2pt]{\scriptsize[$+0.033, +0.061$]}\end{tabular} \\
\llama & \begin{tabular}[c]{@{}c@{}}$+0.068$\\[-2pt]{\scriptsize[$+0.049, +0.086$]}\end{tabular} & 0.48 [0.46] & +0.00 & 8 / +0.03 & $-$0.023 / $-$0.001 & \begin{tabular}[c]{@{}c@{}}$+0.064$\\[-2pt]{\scriptsize[$+0.048, +0.080$]}\end{tabular} \\
\qthree & \begin{tabular}[c]{@{}c@{}}$+0.044$\\[-2pt]{\scriptsize[$+0.029, +0.059$]}\end{tabular} & 0.47 [0.46] & 0.00 & 8 / $-$0.01 & $-$0.011 / +0.004 & \begin{tabular}[c]{@{}c@{}}$+0.037$\\[-2pt]{\scriptsize[$+0.022, +0.051$]}\end{tabular} \\
\bottomrule
\end{tabular}\end{adjustbox}
\caption{Third run with new seeds and failure draws: amendment 10's tests, raw analysis.}
\label{tab:rep3}
\end{table*}

\section{Licenses and compute}\label{app:artifacts}
HotpotQA \citep{yang2018hotpotqa} is distributed under CC BY-SA 4.0 and FEVER \citep{thorne2018fever} under CC BY-SA 3.0 (Wikipedia text, CC BY-SA). \qtwo, \qthree, \qthirty, the Search-R1 framework and the base model of its checkpoint \citep{jin2025searchr1trainingllmsreason} are released under Apache 2.0, and \llama under the Llama 3.1 Community License. \haiku, \sonnet and the other LLM annotators (Claude Fable 5.1, GPT-5.6-luna and GPT-6-sol) were used through their official APIs under the providers' terms. All open models ran locally from their official weights, including the Search-R1 checkpoint released by its authors. Local models ran with vLLM on four NVIDIA RTX A5000 GPUs (24\,GB; \qthirty with tensor parallelism over all four), about 350 GPU-hours in total including development. No model was trained or fine-tuned. Our code, prompts and question manifests are available under the MIT license at \url{https://github.com/bennidict23/judged-useless-queried-anyway}, where the labels will also be released.

\end{document}